\pdfoutput=1

\documentclass[11pt]{article}

\usepackage[final]{acl}

\usepackage{times}
\usepackage{latexsym}
\usepackage{stfloats}

\usepackage[T1]{fontenc}

\usepackage[utf8]{inputenc}

\usepackage{microtype}

\usepackage{inconsolata}

\usepackage{graphicx}
\usepackage{multirow}
\usepackage{makecell} 
\usepackage{tabularx}
\usepackage{graphicx}
\usepackage{booktabs}
\usepackage{geometry}
\usepackage{amsmath}
\usepackage{amsfonts}
\usepackage{amssymb}
\usepackage{float} 
\usepackage{tcolorbox}
\usepackage{xcolor}
\usepackage{enumitem}
\usepackage{pifont}
\usepackage{xspace}
\usepackage{subcaption}
\definecolor{RoyalBlue}{RGB}{65,105,225} 
\definecolor{Green}{RGB}{34,139,34}
\usepackage{fontawesome}

\usepackage[normalem]{ulem}
\newcommand{\ourM}{CoPA} 

\title{\ourM{}: Benchmarking Personalized Question Answering with Data-Informed Cognitive Factors} 

\newcommand{\yinxia}[1]{{\color{blue}{\textsf{[Yingce: #1]}}}}

\author{
 \textbf{Hang Su\textsuperscript{1,2}},
 \textbf{Zequn Liu\textsuperscript{2}\thanks{\xspace Corresponding author.}},
 \textbf{Chen Hu\textsuperscript{2}},
 \textbf{Xuesong Lu\textsuperscript{1}\thanks{\xspace Corresponding author.}},
 \textbf{Yingce Xia\textsuperscript{2}\thanks{\xspace Corresponding author.}},
 \textbf{Zhen Liu\textsuperscript{2}}
\\
 \\
 \textsuperscript{1} East China Normal University, China.
 \textsuperscript{2} Zhongguancun Academy, China.
 \\
 \ttfamily{s-sh25@bza.edu.cn, liuzequn@bza.edu.cn, huchen@bza.edu.cn,}\\
 \ttfamily{xslu@dase.ecnu.edu.cn, xiayingce@bza.edu.cn, pt-lz@bza.edu.cn}
}

\begin{document}
\maketitle
\begin{abstract}
While LLMs have demonstrated remarkable potential in Question Answering (QA), evaluating personalization remains a critical bottleneck. Existing paradigms predominantly rely on lexical-level similarity or manual heuristics, often lacking sufficient data-driven validation. We address this by mining \textit{Community-Individual Preference Divergence} (CIPD), where individual choices override consensus, to distill six key personalization factors as evaluative dimensions. Accordingly, we introduce \ourM{}, a benchmark with 1,985 user profiles for fine-grained, factor-level assessment. By quantifying the alignment between model outputs and user-specific cognitive preferences inferred from interaction patterns, \ourM{} provides a more comprehensive and discriminative standard for evaluating personalized QA than generic metrics. The code is available at \url{https://github.com/bjzgcai/CoPA}.
\end{abstract}

\section{Introduction}

Personalized Question Answering (QA) marks a paradigm shift from static, generic responses to dynamic, user-centric knowledge delivery \cite{nam2009questions,west2014knowledge,lan2022complex,xu2025harnessing}.  For instance, explaining ``gravity'' to a child requires an intuitive narrative involving Newton's apple, whereas a physicist demands a rigorous explanation grounded in mathematical formulas. Recent Large Language Models (LLMs) \citep{achiam2023gpt,team2023gemini,guo2025deepseek,yang2025qwen3} have significantly advanced the field, leveraging their instruction-following and context-adaptation capabilities to dynamically generate such tailored responses. Despite the rapid advancement in LLM-driven personalized QA methods \cite{li2024learning,liu2025one,singh-etal-2024-personal}, establishing a reliable evaluation protocol for personalization remains a critical challenge \cite{salemi2024lamp,kumar2024longlamp,salemi-zamani-2025-lamp}.

\begin{figure}[t!]
\centering
    \includegraphics[width=1.0\columnwidth]{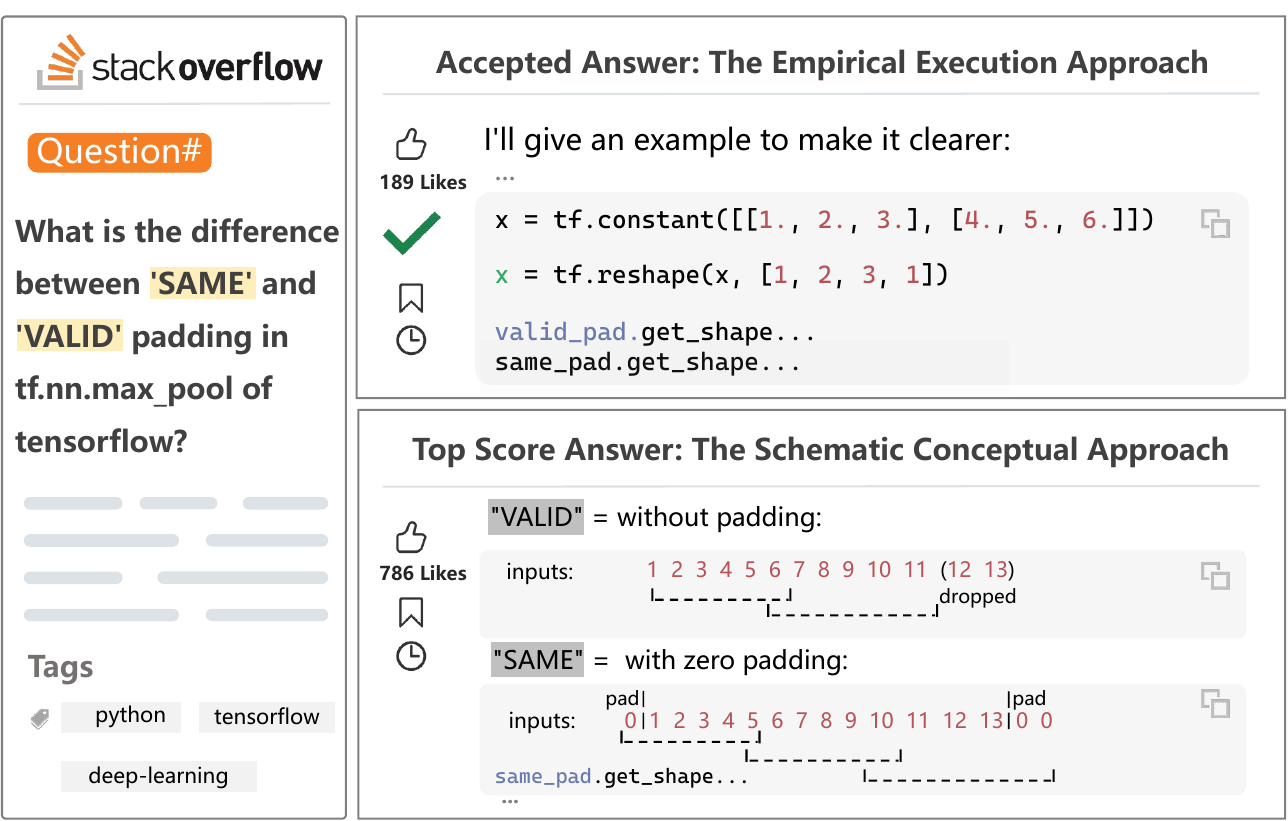}
    \caption{This Stack Overflow example illustrates Community-Individual Preference Divergence (CIPD): the user accepted an empirical, code-based solution while the community overwhelmingly favored a conceptual, schematic one.}
    \label{fig:CIPDQuestion}
\end{figure}


Existing research on the evaluation of personalized generative responses generally falls into three categories. First, traditional NLP metrics \cite{salemi2024lamp,kumar2024longlamp,shi2025retrieval}, such as BLEU \cite{papineni2002bleu} and ROUGE \cite{papineni2002bleu}, focus on the lexical overlap between the generated text and a reference. They fail to capture actual personalized factors, as they only check for matching words rather than adaptation to the user. Second, heuristic-based evaluations~\cite{su-etal-2025-personalized,dai-etal-2024-mpcoder}, attempt to compare the generated response directly to the historical content of users. Although this approach utilizes personalized information, it still relies on lexical-level similarity, failing to capture deeper user intent. Third, the ``LLM-as-a-Judge’’ paradigm~\cite{salemi-zamani-2025-lamp,dong-etal-2024-llm} uses LLMs to assess personalization. 
However, current approaches often rely on generic prompts or manually defined rules to evaluate personalization, lacking the specific criteria or detailed metrics necessary for rigorous assessment. These limitations motivate us to rethink the key factors that determine personalization.

To this end, we adopt a data-driven approach to systematically mine user preferences and disentangle the essential factors of personalization from real-world online QA data. We focus on a prevalent scenario in the online QA systems: Community-Individual Preference Divergence (CIPD) \cite{adamic2008knowledge}. Specifically, in this divergence, the answer endorsed by the community (top-voted) differs from the one adopted by the individual user (accepted), even when both are of high quality (Figure \ref{fig:CIPDQuestion}). Such divergence provides a natural and scalable signal for evaluating personalization.

Leveraging CIPD samples, we distilled personalized factors at scale through the following steps: First, we utilized LLMs to induce user-level decision rationales from CIPD samples. By modeling the user's historical profile alongside the contrasting answer pair, we enabled the LLM to infer the plausible cognitive rationales behind the user's specific choice. We then leveraged LLMs to summarize these rationales, eventually distilling six cognitive factors as evaluative dimensions that govern personalization assessment. 
Notably, although these factors were derived from StackExchange data, we demonstrate in this work that the resulting factor-based evaluation exhibit robust generalization capabilities when applied to new scenarios. 

Based on these data-informed factors, we constructed a personalized QA benchmark, CoPA, to evaluate personalized question answering with LLMs. The benchmark comprises a curated set of 1,985 users from diverse domains, each associated with a historical interaction profile. For evaluation, we quantified user profiles across the six factors and adopted a 3-point Likert scale to measure the alignment between model-generated responses and these profiles. 


Our contribution can be summarized as follows:

\noindent(1) We utilized LLMs to automatically conduct an in-depth exploration of the CIPD phenomenon at the user level and distilled a set of interpretable cognitive factors from large-scale online QA data.
   
\noindent(2) We propose \ourM{}, a novel factor-driven benchmark that enables quantitative evaluation of alignment between personalized content generated by LLMs and users' cognitive preferences as reflected in interaction patterns.

\noindent(3) We conducted extensive experiments on the \ourM{} benchmark, established comprehensive baselines, and provided in-depth quantitative and qualitative analyses, offering a solid empirical foundation for future research.

\section{Analysis of CIPD Phenomenon}
We investigate the CIPD phenomenon by leveraging data from StackExchange\footnote{\url{https://stackexchange.com/}}, a representative multi-domain Q\&A platform. 

\textbf{StackExchange dataset.}
Stack Exchange serves as a massive online Q\&A network encompassing 173 distinct communities, characterized by a diverse user base of over 100 million monthly visitors. These domains can be categorized into four primary groups: \textbf{Engineering \& Tools}, \textbf{Science \& Theory}, \textbf{Lifestyle \& Society}, and \textbf{Leisure \& Fandom}, comprehensively covering the majority of real-world QA scenarios. Table \ref{tab:Distribution of domains and question counts.} presents the top four domains within each category based on question volume, along with their respective distributions. On this platform, a single question elicits multiple responses, and the community votes to reflect the general consensus. Concurrently, the question asker selects a specific response using an ``accepted'' label to indicate his/her personal preference. This explicit distinction between community endorsement and individual choice renders Stack Exchange an ideal environment for investigating the CIPD phenomenon. A comprehensive statistical breakdown of question counts for all domains is provided in the Appendix \ref{sec:Question Distribution}.

\begin{table}[htbp]
    \centering
    \small  
    \renewcommand{\arraystretch}{0.9} 
    \setlength{\tabcolsep}{4pt}      
    \begin{tabular}{llrr}
        \toprule
        \textbf{Category} & \textbf{Domain} & \textbf{Count} & \textbf{Pct.(\%)} \\
        \midrule
        \multirow{5}{*}{\makecell[l]{\textbf{Engineering \& Tools} \\ \scriptsize{(Total: 1,537,431)}}} 
          & unix & 158,149 & 10.3 \\
          & tex & 119,815 & 7.8 \\
          & salesforce & 85,487 & 5.6 \\
          & electronics & 81,643 & 5.3 \\
          & \textit{Other} & 1,092,337 & 71.0 \\
        \midrule
        
        \multirow{5}{*}{\makecell[l]{\textbf{Science \& Theory} \\ \scriptsize{(Total: 1,652,195)}}} 
          & math & 1,109,289 & 67.1 \\
          & physics & 136,393 & 8.3 \\
          & stats & 104,284 & 6.3 \\
          & mathematica & 50,863 & 3.1 \\
          & \textit{Other} & 251,366 & 15.2 \\
        \midrule
        
        \multirow{5}{*}{\makecell[l]{\textbf{Lifestyle \& Society} \\ \scriptsize{(Total: 664,689)}}} 
          & english & 96,299 & 14.5 \\
          & ell & 79,979 & 12.0 \\
          & diy & 44,782 & 6.7 \\
          & money & 28,599 & 4.3 \\
          & \textit{Other} & 415,031 & 62.4 \\
        \midrule
        
        \multirow{5}{*}{\makecell[l]{\textbf{Leisure \& Fandom} \\ \scriptsize{(Total: 287,908)}}} 
          & gaming & 77,043 & 26.8 \\
          & scifi & 53,021 & 18.4 \\
          & travel & 33,644 & 11.7 \\
          & rpg & 31,922 & 11.1 \\
          & \textit{Other} & 92,301 & 32.1 \\
        \bottomrule
    \end{tabular}
    \caption{Distribution of domains and question counts.}
    \label{tab:Distribution of domains and question counts.}
\end{table}

\begin{figure}[tbp]
    \centering
    \begin{subfigure}[b]{0.9\columnwidth}
        \includegraphics[width=0.9\columnwidth]{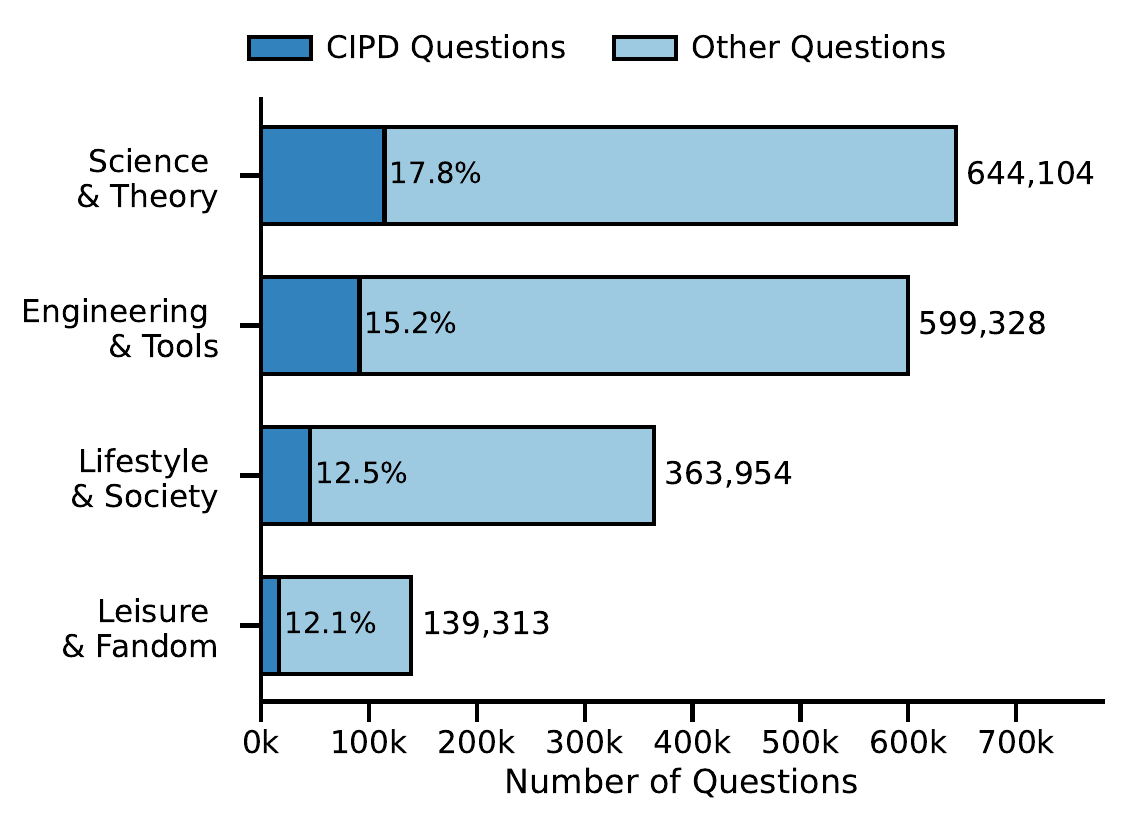}
        \caption{Proportion of CIPD questions.}
        \label{fig:CIPDquestion_Proportion}
    \end{subfigure}
    
    
    \begin{subfigure}[b]{0.9\columnwidth}
        \centering
        \includegraphics[width=0.9\columnwidth]{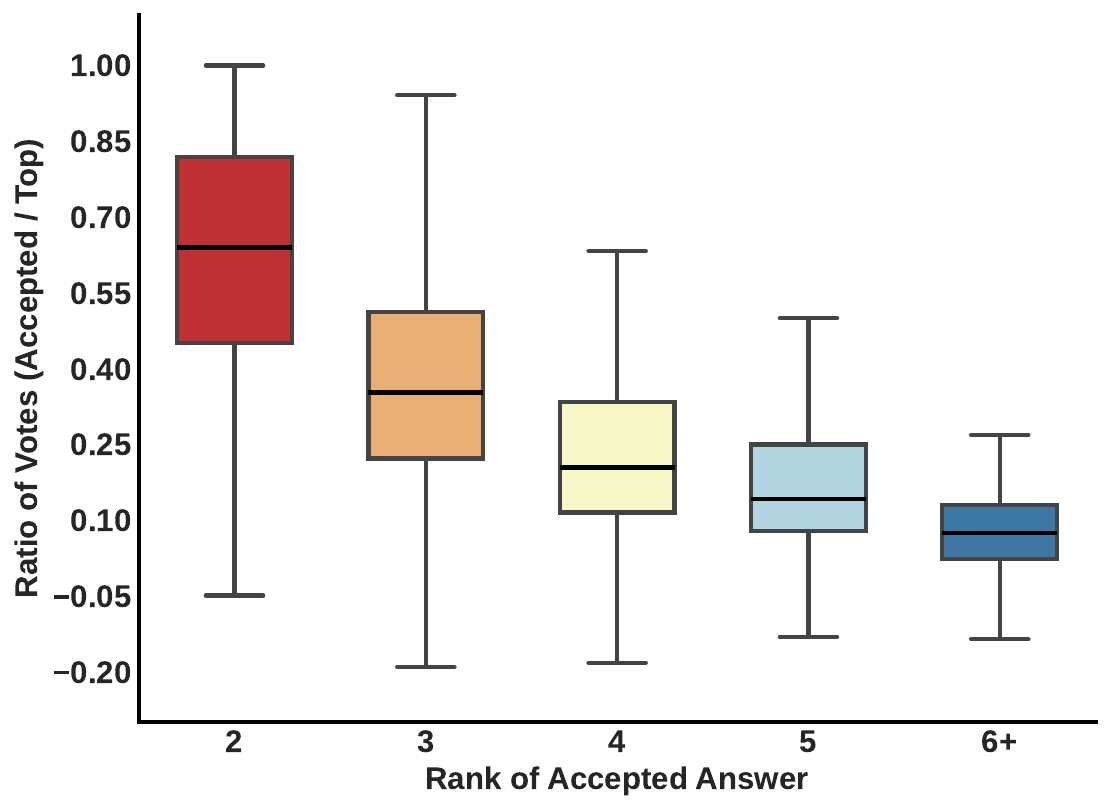}
        \caption{Rank distribution of CIPD questions.}
        \label{fig:CIPDquestion_rank}
    \end{subfigure}
    \caption{Analysis of CIPD questions. (a) shows the proportion across categories, while (b) illustrates the rank distribution.}
    \label{fig:CIPD_combined}
\end{figure}

\textbf{CIPD Questions.}
In this analysis, we exclude questions containing only a single answer. As illustrated in the Figure \ref{fig:CIPDquestion_Proportion}, the CIPD phenomenon is pervasive across all primary categories, notably persisting even in domains typically assumed to have objective factual answers, such as Engineering \& Tools. Figure \ref{fig:CIPDquestion_rank} reveals significant disparities in vote counts between accepted and top-voted answers when divergence occurs, indicating that individual users frequently prioritize distinct criteria over community consensus. 
To examine whether answer timing affects CIPD, we conduct a statistical analysis of answer timestamps for CIPD questions, showing that 72.70\% of accepted answers were posted no earlier than the corresponding top-voted answers.
Overall, the results confirm the prevalence of the CIPD phenomenon within the StackExchange dataset, reflecting the personalized needs of individual users.

\section{Factor distillation}

\begin{figure*}[tbp]
\centering
    \includegraphics[width=2\columnwidth]{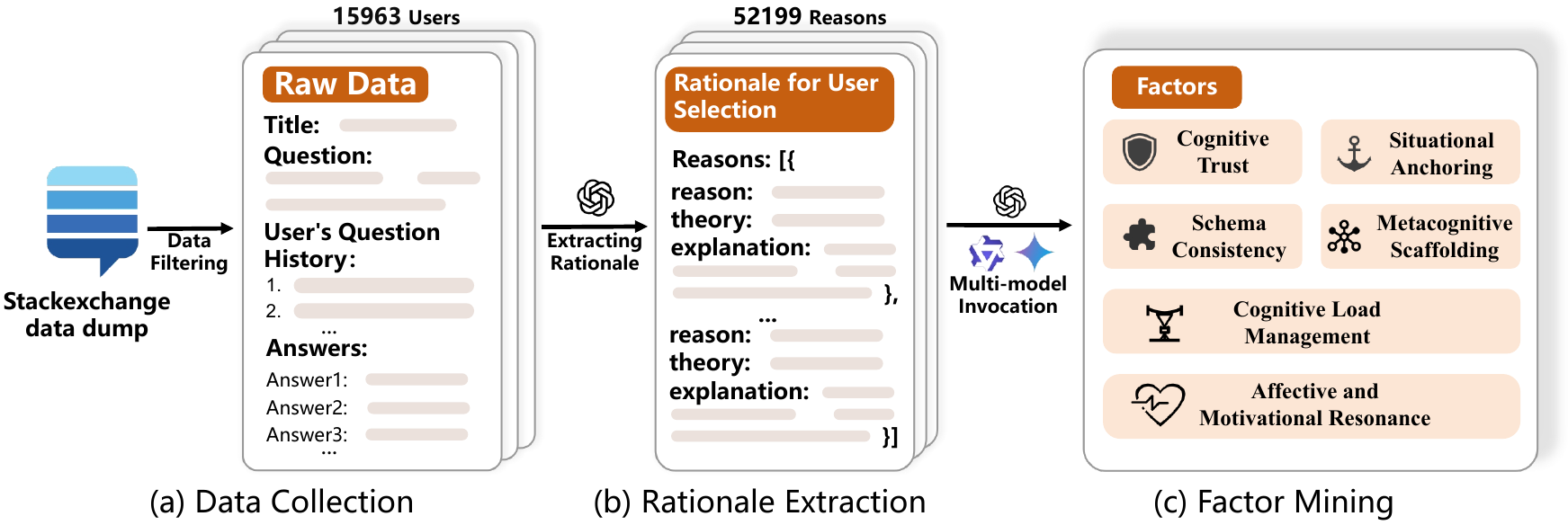}
    \caption{The pipeline of the Factor Distillation process. We first collect and filter raw CIPD data (a), then employ LLMs to extract decision rationales behind user choices (b), and finally distill these rationales into six cognitive factors through multi-model consensus (c).}
    \label{fig:dataflow}
\end{figure*}

This section outlines our method for mining personalized factors from the StackExchange dataset. Figure~\ref{fig:dataflow} shows the pipeline of the Factor Distillation process. First, we curated a high-quality CIPD dataset by linking user QA histories with community voting information. Second, we leveraged LLMs to extract the post-hoc decision rationales behind user choices. Finally, we aggregated these rationales to distill core factors, which were subsequently verified through expert validation. 

{\em Data Curation}: The data curation pipeline consists of a series of cascaded operations. We filtered out samples containing hyperlinks, excessive text length with more than 4096 words, and zero answers; these steps eliminated 1,965,220, 5,310, and 734,011 instances, respectively. To focus on users exhibiting personalized preferences, we selected only those who posted at least three CIPD questions across all domains. From an initial pool of 6.8 million StackExchange questions, the final dataset contains 626,786 questions grouped by 15,963 users.

{\em Notations}: define our dataset as $\mathcal{U} = \{U_1, U_2, \dots, U_N\}$, where $N$ represents the total number of users ($N=15,963$). Each  $U_i$ is associated with a sequence of data instances $(u_{i,1}, u_{i,2}, \dots, u_{i,T_i})$, where $T_i$ denotes the number of tracked QA pairs for user $i$. The instance $u_{i,j}$ for any $j\in\{1,2,\cdots,T_i\}$ is defined as a tuple:
\begin{equation}
u_{i,j} = \left(q_{i,j}, (a_{i,j}^*, v_{i,j}^*), \left\{(a_{i,j}^{(k)}, v_{i,j}^{(k)})\right\}_{k=1}^{N_{i,j}}\right).
\label{eq:user_ij_data_example}
\end{equation}
In Eqn. \eqref{eq:user_ij_data_example}, $q_{i,j}$ denotes the $j$-th question of user $i$. The pair $(a_{i,j}^*, v_{i,j}^*)$ represents the user-selected (accepted) answer and its corresponding vote count.  Similarly, $(a_{i,j}^{(k)}, v_{i,j}^{(k)})$ represents the $k$-th candidate answer along with its vote count, where $N_{i,j}$ denotes the total number of collected answers for question $q_{i,j}$. 
We define $(a_{i,j}^t, v_{i,j}^{t})$ as the answer with the maximal vote count and that specific vote count, respectively. In our CIPD dataset, $a_{i,j}^t \neq a_{i,j}^*$ and $v_{i,j}^t > v_{i,j}^*$.

{\em Rationale extraction}: 
We employed an LLM to extract personalized rationales from the QA trajectory of user $i$. To ensure computational efficiency, we input each instance $u_{i,j}$, comprising the current question ($q_{i,j}$), the explicitly marked user-selected answer ($a^*$), and the remaining answers $a_{i,j}^{(k)}$ for any $k\in [N_{i,j}]$, along with a context of at most recent 50 historical questions ($q_{i,j-49},\cdots,q_{i,j}$), into the LLM, which was then prompted to deduce the plausible cognitive drives behind the user's choices. Specifically, we enforced a structured JSON output schema that steers the generation through explicit reasoning steps:  identifying a behavioral label (``reason''), mapping it to a descriptive psychological framework (``theory''), and synthesizing a justification for this alignment (``explanation'').
To validate our approach, we randomly sampled 500 instances and conducted a manual inspection to assess the reliability of the generated explanations. The specific evaluation criteria are detailed in Appendix \ref{sec:Manual Inspection Criteria for Reason Analysis}. We invited two domain experts to perform this evaluation, obtaining a Cohen's Kappa of 0.7251 and a 97\% alignment rate with expert judgment, which together support the reliability of our extraction pipeline.


{\em Factor Distilling}:
Building upon the aforementioned relationships, we aggregated the discrete rationales into a candidate pool and employed LLMs to distill core influencing factors. The prompt is shown in Figure~\ref{fig:Factor Mine Prompt} of Appendix~\ref{sec:Factor Mining Prompt}. To ensure the stability and robustness of the extraction results, we adopted an ensemble strategy involving multiple models and repeated invocations, selecting factors with the highest consensus via frequency statistics. Subsequently, to conduct a qualitative plausibility assessment, two expert volunteers with Ph. D.s in Psychology were invited to manually evaluate and verify the results. The six distinct factors identified through this process are:
(1) \textbf{Cognitive Trust} (CT): Does the response align with the user's threshold for trust and credibility within this specific domain? (2) \textbf{Situational Anchoring} (SA): Is the response precisely calibrated to the user's immediate context and specific problem? (3) \textbf{Schema Consistency} (SC): Does the response integrate coherently with the user's prior knowledge and existing mental models? (4) \textbf{Cognitive Load Management} (CLM): Is the complexity of the response tailored to match the user's cognitive capacity? (5) \textbf{Metacognitive Scaffolding} (MS): Does the response provide structural support that fosters the user's critical thinking skills? (6) \textbf{Affective and Motivational Resonance} (AMR): Does the response resonate with the user's current emotional state and motivational orientation? Three specific cases and the expert evaluation criteria are provided in Appendix \ref{sec:Factors and Their Associated Causal Cases}.

\section{The \ourM{} Benchmark}\label{sec:The CoPA Benchmark}
We establish the \ourM{} benchmark to rigorously evaluate personalization methods using the six core factors. These factors serve as fine-grained dimensions for assessing the alignment between model-generated content and user cognitive needs.

\subsection{Benchmark Construction}
\noindent\textbf{Data collection}. 
We constructed the evaluation set by sampling the top 20\% of users per domain, ranked by the vote counts of their most recent questions, to ensure the data quality. Domains with fewer than 10 users were fully included, while domains with more than 50 candidates were restricted to 50 users. This process yielded 1,985 unique users. Finally, these users and their corresponding interaction histories constitute our \ourM{} benchmark.
\paragraph{User Profile Construction.}
For each user $i$ for any $i\in[N]$ in the benchmark, we constructed the user preference profile, denoted as $P_i$, across the six identified factors. By feeding an LLM with the user's 50 recent interactions, we prompted the model to quantify the user's preference for each factor $f_k \in \{$CT, SA, SC, CLM, MS, AMR$\}$. This process yielded a structured profile $P_i = \{p_1, p_2, ..., p_6\}$, where $p_k$ represents the user's preference on factor $f_k$.

Table~\ref{tab:Dataset statistics} presents the detailed statistics of the \ourM{} benchmark constructed via this pipeline.

\begin{figure}[t]
\centering
    \includegraphics[width=1\columnwidth]{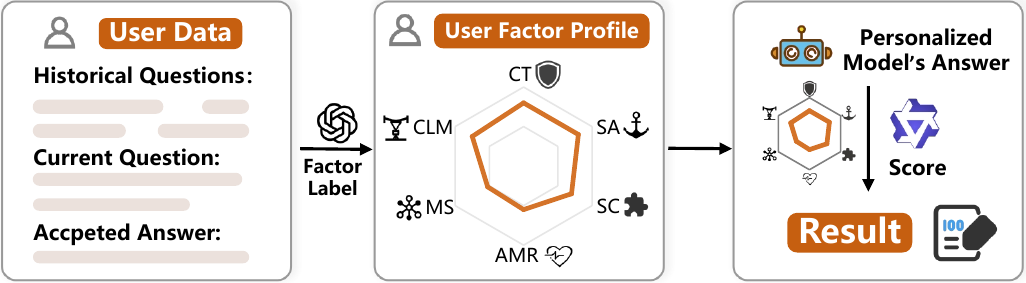}
    \caption{The proposed evaluation pipeline.}
    \label{fig:Evaluation Metric}
\end{figure}

\paragraph{Evaluation Metric.}
We evaluate the response $R_i$ generated for user $i$ by measuring its alignment with the established profile $P_i$. An evaluator LLM compares $R_i$ against each $p_k \in P_i$ to assign an alignment score $s_k$ for any $k\in\{1,2,\cdots,6\}$. Prompt details are left in Appendix~\ref{sec:Factor Label Prompt}, Figure~\ref{fig:Evaluation Prompt system}. We adopt a 3-point Likert scale, as commonly used in recent LLM-based evaluation studies \cite{baek2024knowledge,salemi-zamani-2025-lamp}, to quantify the alignment score $s_k$:
\begin{itemize}[leftmargin=*, noitemsep, topsep=0pt]
    \item 0 (Mismatch): The response violates or ignores user preferences.
    \item 1 (Partial Match): The response aligns partially but lacks sufficient depth.
    \item 2 (Full Match): The response perfectly adapts to the user's factor profile.
\end{itemize}
The final scores are normalized to the range $[0, 1]$ for aggregation. Figure \ref{fig:Evaluation Metric} demonstrates the user profile construction and evaluation process.


\begin{table}[t] 
\centering
\label{tab:stats}
\resizebox{\columnwidth}{!}{%
\renewcommand{\arraystretch}{1.2} 
\begin{tabular}{lcccc}
\toprule
\textbf{Statistics} & \textbf{\begin{tabular}[c]{@{}c@{}}Eng.\\ Tools\end{tabular}} & \textbf{\begin{tabular}[c]{@{}c@{}}Sci.\\ Theory\end{tabular}} & \textbf{\begin{tabular}[c]{@{}c@{}}Life.\\ Society\end{tabular}} & \textbf{\begin{tabular}[c]{@{}c@{}}Leisure\\ Fandom\end{tabular}} \\ 
\midrule
\textbf{Users} & 864 & 456 & 413 & 252 \\
\textbf{Records (Avg. Q/User)} & 30.69 & 27.57 & 41.09 & 40.74 \\
\addlinespace 
\textbf{Q. Title Length} & 8.96 & 9.92 & 9.64 & 9.69 \\
\textbf{Q. Body Length} & 128.21 & 131.23 & 109.49 & 135.85 \\
\addlinespace
\textbf{Factors Profile Size} & 555.12 & 597.74 & 580.34 & 590.10 \\ 
\bottomrule
\end{tabular}%
}
\caption{Dataset statistics of the \ourM{} benchmark.}
\label{tab:Dataset statistics}
\end{table}

\subsection{Evaluated Approaches}
We evaluated four representative approaches for personalized QA on our benchmark.

\noindent\textbf{No-Personalization}  generates responses directly using the LLM without access to user profiles. It constitutes a general-purpose, non-personalized baseline. The prompt is shown in Figure~\ref{fig:No-Personalization Prompt} of Appendix~\ref{sec:Baselines for the CoPA Benchmark Prompt}.

\noindent\textbf{Time-Personalization} \citep{cattell1963theory} assumes that a user's cognitive state evolves over time. We construct the user profile using the $K$ most recent historical questions in chronological order to prompt the LLM. The prompt is in Figure~\ref{fig:Time-Personalization Prompt}.

\noindent\textbf{RAG-Personalization} \citep{salemi2024lamp} retrieves the top-$K$ historical questions with the highest semantic affinity to the current query. By aggregating these relevant instances, the model infers the user's specific attentional focus to generate tailored responses. The prompt is shown in Figure~\ref{fig:RAG-Personalization Prompt} of Appendix~\ref{sec:Baselines for the CoPA Benchmark Prompt}.

\noindent\textbf{Profile-Personalization} \citep{su-etal-2025-personalized} compresses the user's sequence of $K$ recent questions into Domain Profiles and Global Profiles, thereby extracting higher-dimensional personalized information. See Appendix \ref{sec:Profile-Personalization Prompt} for details.
\section{Experiments}

\subsection{Experimental Setup}
For the rationale extraction task, we employ the GPT-5-chat-latest model with the temperature parameter set to 0.7. In the Factor Mining phase, we ensembled the results from GPT-5-chat-latest\footnote{\url{https://chatgpt.com/}}, Gemini-2.5-Pro\footnote{\url{https://gemini.google.com/}}, and Qwen3-Max\footnote{\url{https://chat.qwen.ai/}}, all configured with a temperature of 0.7. Regarding the four approaches for personalized QA, we select the Qwen3-8B (Fast-Thinking mode) \cite{yang2025qwen3} and the GPT-4o-mini \cite{hurst2024gpt} to serve as backbone models. Furthermore, within the RAG personalization module, Qwen3-Embedding-4B \cite{zhang2025qwen3} is adopted as the embedding model. For evaluation, we use Qwen3-32B (Fast-Thinking mode) \cite{yang2025qwen3}. In terms of hardware and deployment, all local models are hosted on a computing platform equipped with four NVIDIA A800 GPUs, utilizing the vLLM\footnote{\url{https://docs.vllm.ai/en/stable/}} framework for inference acceleration.

\subsection{Effectiveness of Factors}\label{sec:compare with top voted}
We conducted a quantitative evaluation on \ourM{} to validate the alignment of our factor-based scoring with ground-truth user preferences. We utilized the factor profile constructed from the preceding CIPD question to evaluate the accepted and top-voted answers, obtaining scores denoted as $S_{acc}$ and $S_{top}$, respectively. We defined three evaluation metrics: \textbf{Accuracy}: The proportion of instances where $S_{acc} > S_{top}$. \textbf{Tie Rate}: The proportion of instances where $S_{acc} = S_{top}$. \textbf{Score Margin}: The difference between $S_{acc}$ and $S_{top}$, calculated as $S_{acc} - S_{top}$. 
An evaluation metric that effectively captures user preference should maximize Accuracy and Margin while minimizing the Tie Rate, thereby demonstrating a stronger capability to distinguish the user's personal choice from the community consensus.
We compared the factor-based scoring with four evaluation methods: (1) Direct scoring via LLM-as-a-Judge on a 3-point Likert scale; (2) CoT scoring via LLM-as-a-Judge with few-shot demonstrations and explicit reasoning steps on a 3-point Likert scale; (3) heuristic-based evaluation metrics (Jaccard and Inclusion Coefficient) derived from \citep{su-etal-2025-personalized} (these two metrics are not normalized; therefore, the Tie Rate and Score Margin are not calculated.); (4) scoring using randomly generated factors by GPT-5-chat-latest, adapted from the methodology in \citep{salemi2024lamp}. Crucially, to prevent data leakage, all data samples in \ourM{} benchmark were truncated to the preceding CIPD question for factor construction. The prompts for the Direct, CoT, and Random baselines are provided in Appendix~\ref{sec:Effectiveness of Factors}.

The results are presented in Table \ref{tab:Comparison of Evaluation Methodologies}. 
Distinguishing ``Accepted'' from ``Top-voted'' answers proves to be a non-trivial challenge. Top-voted answers often feature high quality, serving as potent distractors for the evaluator LLM. This is evident in the Direct baseline, which exhibits excessive Tie Rates and low Accuracy. Although CoT scoring improves over Direct scoring, it still leaves Tie Rates above 55\% in all domains and Accuracy below 32\%, indicating that more elaborate prompting alone is insufficient. 
Compared to Direct and CoT scoring, our evaluation based on distilled factors consistently reduces Tie Rates to 18.24\%--26.28\% and improves Accuracy to 51.43\%--55.37\%, demonstrating that distilled factors enable the identification of granular features aligned with user intent. Relative to heuristic metrics (Jaccard/Inclusion), which perform better than Direct scoring but rely on surface-level lexical overlap, our method maintains a clear lead by capturing deeper cognitive alignment. Against the baseline with randomly generated factors, distilled factors still yield sizable gains across all domains, confirming the effectiveness of data-driven factor distilling rather than prompt sophistication alone. For completeness, the corresponding results when the factor profile is constructed from the current CIPD question are reported in Appendix~\ref{sec:Effectiveness of Factors}, Table~\ref{tab:Comparison of Evaluation Methodologies_appendix}. 

To investigate the effectiveness of each factor, we conducted an ablation study (Table \ref{tab:ablation}). The results indicate that removing any individual factor leads to a degradation in performance in terms of Accuracy, Tie Rate, and Score Margin, thereby demonstrating the necessity of each factor in capturing the user preferences.

\begin{table*}[t!]
\centering
\small
\renewcommand{\arraystretch}{1.2} 
\setlength{\tabcolsep}{0pt} 
\begin{tabular*}{\textwidth}{@{\extracolsep{\fill}} l ccc ccc ccc ccc @{}}
\toprule
\multirow{2}{*}{\textbf{Method}} & \multicolumn{3}{c}{\textbf{Eng. \& Tools}} & \multicolumn{3}{c}{\textbf{Science}} & \multicolumn{3}{c}{\textbf{Lifestyle}} & \multicolumn{3}{c}{\textbf{Leisure}} \\
\cmidrule(lr){2-4} \cmidrule(lr){5-7} \cmidrule(lr){8-10} \cmidrule(lr){11-13}
 & Acc$\uparrow$ & Tie$\downarrow$ & Mar.$\uparrow$ & Acc$\uparrow$ & Tie$\downarrow$ & Mar.$\uparrow$ & Acc$\uparrow$ & Tie$\downarrow$ & Mar.$\uparrow$ & Acc$\uparrow$ & Tie$\downarrow$ & Mar.$\uparrow$ \\
\midrule
Direct       & 12.15 & 80.67 & 0.026 & 18.86 & 75.00 & 0.065 & 10.17 & 84.75 & 0.027 & 11.90 & 81.75 & 0.028 \\
CoT          & 26.00 & 60.31 & 0.061 & 31.35 & 55.16 & 0.091 & 26.39 & 61.74 & 0.079 & 29.55 & 59.85 & 0.106 \\
Jaccard      & 40.28 & --    & --    & 42.32 & --    & --    & 39.23 & --    & --    & 42.06 & --    & --    \\
Incl. Coeff. & 34.07 & --    & --    & 26.54 & --    & --    & 27.60 & --    & --    & 28.17 & --    & --    \\
Random       & 25.58 & 61.00 & 0.049 & 30.26 & 56.14 & 0.074 & 33.90 & 53.27 & 0.081 & 35.09 & 50.00 & 0.071 \\
\textbf{ours} & \textbf{51.57} & \textbf{26.28} & \textbf{0.127} & \textbf{53.44} & \textbf{22.88} & \textbf{0.136} & \textbf{51.43} & \textbf{24.12} & \textbf{0.093} & \textbf{55.37} & \textbf{18.24} & \textbf{0.143} \\
\bottomrule
\end{tabular*}
\caption{Comparison of Evaluation Methodologies. Arrows ($\uparrow$/$\downarrow$) indicate whether higher or lower values are better. Values are in percentages (\%) except for Margin.}
\label{tab:Comparison of Evaluation Methodologies}
\end{table*}

\begin{table*}[t!]
\centering
\small 
\renewcommand{\arraystretch}{1.2} 
\setlength{\tabcolsep}{0pt} 

\begin{tabular*}{\textwidth}{@{\extracolsep{\fill}} l ccc ccc ccc ccc @{}}
\toprule
\multirow{2}{*}{\textbf{Factors}} & \multicolumn{3}{c}{\textbf{Eng. \& Tools}} & \multicolumn{3}{c}{\textbf{Science}} & \multicolumn{3}{c}{\textbf{Lifestyle}} & \multicolumn{3}{c}{\textbf{Leisure}} \\
\cmidrule(lr){2-4} \cmidrule(lr){5-7} \cmidrule(lr){8-10} \cmidrule(lr){11-13}
 & Acc$\uparrow$ & Tie$\downarrow$ & Mar.$\uparrow$ & Acc$\uparrow$ & Tie$\downarrow$ & Mar.$\uparrow$ & Acc$\uparrow$ & Tie$\downarrow$ & Mar.$\uparrow$ & Acc$\uparrow$ & Tie$\downarrow$ & Mar.$\uparrow$ \\
\midrule
w/o CT   & 54.28 & 29.40 & 0.177 & 54.82 & 31.36 & 0.203 & 54.96 & 26.15 & 0.172 & 54.76 & 26.19 & 0.198 \\
w/o SA   & 54.05 & 31.02 & 0.182 & 52.41 & 33.99 & 0.208 & 55.21 & 25.91 & 0.173 & 55.95 & 26.59 & 0.208 \\
w/o SC   & 54.63 & 29.17 & 0.181 & 54.17 & 32.02 & 0.206 & 54.72 & 26.39 & 0.174 & 56.35 & 26.19 & 0.207 \\
w/o CLM  & 53.24 & 31.37 & 0.175 & 53.95 & 32.68 & 0.206 & 55.69 & 24.94 & 0.171 & 57.14 & 26.19 & 0.216 \\
w/o MS   & 50.00 & 34.61 & 0.174 & 51.54 & 35.96 & 0.199 & 53.51 & 28.57 & 0.170 & 53.57 & 28.17 & 0.200 \\
w/o AM   & 53.82 & 30.21 & 0.181 & 53.73 & 33.11 & 0.203 & 55.21 & 26.88 & 0.174 & 55.56 & 26.19 & 0.208 \\
\textbf{All} & \textbf{55.09} & \textbf{28.24} & \textbf{0.189} & \textbf{55.04} & \textbf{30.92} & \textbf{0.211} & \textbf{56.17} & \textbf{24.21} & \textbf{0.180} & \textbf{57.14} & \textbf{24.21} & \textbf{0.216} \\
\bottomrule
\end{tabular*}
\caption{Ablation Study of Different Factors. Arrows ($\uparrow$/$\downarrow$) indicate whether higher or lower values are better. Values are in percentages (\%) except for Margin.}
\label{tab:ablation}
\end{table*}

\subsection{Generalizability of the Factors}
To validate the generalizability of the mined factors, we extended our evaluation to two external benchmark datasets: UPGC-QA \cite{su-etal-2025-personalized} and LaMP-QA \cite{salemi-zamani-2025-lamp}. 
We sought to evaluate whether factors can enhance the performance of baselines on these benchmarks. Specifically, we integrated our factors into the system prompts of the baseline methods, guiding the models to focus on these dimensions during generation. As shown in Table~\ref{tab:lampqa} and Table~\ref{tab:UPGCQA} (the prompt and result  are detailed in Appendix \ref{sec:Generalizability of the Factors_appendix}.), incorporating these factors, yielded consistent and significant performance gains across two distinct benchmark, two different backbone models, and all evaluated personalization methods. This improvement indicates the factors' generalizability, suggesting that their optimization is a promising direction for personalized QA and thereby motivating the use of our factor-based benchmark, \ourM{}.
\begin{table}[t]
\centering
\small
\renewcommand{\arraystretch}{1.2}
\setlength{\tabcolsep}{3pt} 
\begin{tabular}{l cc cc cc}
\toprule
\multirow{2}{*}{\textbf{Method}} & \multicolumn{2}{c}{\textbf{Wildchat}} & \multicolumn{2}{c}{\textbf{StackEx.}} & \multicolumn{2}{c}{\textbf{CS101}} \\
\cmidrule(lr){2-3} \cmidrule(lr){4-5} \cmidrule(lr){6-7}
 & \textbf{Jac.} & \textbf{I.C.} & \textbf{Jac.} & \textbf{I.C.} & \textbf{Jac.} & \textbf{I.C.} \\
\midrule
\multicolumn{7}{c}{\textbf{Qwen3 8B}} \\
\midrule
DirectQA      & 6.52 & 8.29 & 40.64 & 64.86 & 15.56 & 42.86 \\
RAGQA         & 6.57 & 8.64 & 40.72 & 65.22 & 16.59 & 43.02 \\
\hspace{2mm}\textit{w/ Factors} & \textbf{7.88} & \textbf{9.33} & \textbf{41.99} & \textbf{66.71} & \textbf{17.15} & \textbf{43.96} \\
ProfileQA     & 6.99 & 8.87 & 43.01 & 73.02 & 16.32 & 45.99 \\
\hspace{2mm}\textit{w/ Factors} & \textbf{8.31} & \textbf{10.03} & \textbf{44.17} & \textbf{74.23} & \textbf{17.67} & \textbf{47.35} \\
\midrule
\multicolumn{7}{c}{\textbf{GPT-4o-mini}} \\
\midrule
DirectQA      & 7.11 & 9.27 & 42.33 & 64.79 & 15.77 & 43.13 \\
RAGQA         & 7.86 & 9.89 & 43.34 & 65.23 & 16.94 & 43.21 \\
\hspace{2mm}\textit{w/ Factors} & \textbf{8.31} & \textbf{10.22} & \textbf{44.57} & \textbf{67.14} & \textbf{17.11} & \textbf{43.99} \\
ProfileQA     & 8.22 & 11.35 & 44.09 & 72.98 & 16.37 & 46.23 \\
\hspace{2mm}\textit{w/ Factors} & \textbf{9.64} & \textbf{11.67} & \textbf{44.96} & \textbf{75.05} & \textbf{18.24} & \textbf{47.98} \\
\bottomrule
\end{tabular}
\caption{Comparison of personalization methods with and without factor augmentation on three datasets. \textbf{Jac.}: Jaccard, \textbf{I.C.}: Incl. Coeff.}
\label{tab:UPGCQA}
\end{table}

\subsection{Factor-level evaluation of personalized QA approaches using \ourM{}}
Table \ref{tab:baselines} presents a comparative evaluation of Time-, RAG-, and Profile-Personalization methods against the non-personalized baseline and ground truth across four categories. 

The results demonstrated that incorporating personalized context consistently yields performance gains in general. 
Among the three personalization strategies, Profile-Personalization demonstrated the most superior performance on all four sub-domains. This suggests that constructing structured and semantically rich user profiles allows for a more precise capture of user intent, thereby generating responses that are better aligned with user needs. Additionally, the performances of RAG-Personalization and Time-Personalization surpassed the top-voted answer, indicating that leveraging retrieval augmentation and temporal information serves as an effective personalization mechanism.

Regarding the comparison of base models, Qwen 3 8B outperformed GPT-4o-mini overall on this benchmark. Even in the No-Personalization setting, the average score of Qwen 3 8B surpassed that of GPT-4o-mini. When combined with Profile-Personalization, Qwen 3 8B further narrowed the gap with human-accepted answers, demonstrating the model's strong potential in handling personalized question-answering tasks.

From a category perspective, models generally excelled in the Engineering \& Tools but underperformed in Science \& Theory. We attribute this disparity to the latter's inherent demand for deep reasoning and specialized domain knowledge, which challenges current personalization models. This finding highlights a critical avenue for future research: enhancing the capability of personalized systems to navigate complex theoretical and high-cognitive-load scenarios.\\
Our detailed factor-level analysis (Appendix \ref{sec:Detailed factor-level analysis}) reveals several key insights. The inclusion of user profiles yielded substantial gains across Cognitive Trust, Schema Consistency, and Situational Anchoring. This indicates that explicit user modeling enhances contextual alignment and fosters deeper cognitive trust. In stark contrast, Metacognitive Scaffolding remains a significant challenge, with all models scoring exceptionally low, highlighting the inherent difficulty of inducing user reflection in generative tasks. Conversely, Cognitive Load Management scores were consistently high and were further improved by personalization, demonstrating efficacy in tailoring information density. Finally, marked improvements in Affective and Motivational Resonance confirm that profile-aware generation moves beyond mere information delivery to achieve genuine user resonance by capturing emotional nuances.

\begin{table}[t]
\centering
\small 
\renewcommand{\arraystretch}{1.2}
\setlength{\tabcolsep}{2.5pt} 
\begin{tabular}{@{} l ccccc @{}}
\toprule
\textbf{Method} & \textbf{E\&T} & \textbf{S\&T} & \textbf{L\&S} & \textbf{L\&F} & \textbf{Avg.} \\
\midrule
Accepted Ans.   & \textbf{0.8009} & \textbf{0.7984} & \textbf{0.7393} & \textbf{0.7946} & \textbf{0.7833} \\
Random Ans.     & 0.6183 & 0.5774 & 0.5516 & 0.5418 & 0.5722 \\
Top Ans.        & 0.6207 & 0.5941 & 0.5633 & 0.5886 & 0.5917 \\
\midrule
\multicolumn{6}{c}{\textbf{GPT-4o-mini}} \\
\midrule
No-Pers.        & 0.6307 & 0.4700 & 0.5347 & 0.5136 & 0.5373 \\
Time-Pers.      & 0.6858 & 0.5353 & 0.5591 & 0.5516 & 0.5830 \\
RAG-Pers.       & 0.6964 & 0.5334 & 0.5748 & 0.5472 & 0.5909 \\
Profile-Pers.   & \textbf{0.7003} & \textbf{0.5912} & \textbf{0.6223} & \textbf{0.5719} & \textbf{0.6214} \\
\midrule
\multicolumn{6}{c}{\textbf{Qwen 3 8B}} \\
\midrule
No-Pers.        & 0.6671 & 0.5409 & 0.5803 & 0.5403 & 0.5821 \\
Time-Pers.      & 0.7188 & 0.5691 & 0.6230 & 0.5638 & 0.6187 \\
RAG-Pers.       & 0.7198 & 0.5724 & 0.6172 & 0.5721 & 0.6204 \\
Profile-Pers.   & \textbf{0.7341} & \textbf{0.6365} & \textbf{0.6729} & \textbf{0.6002} & \textbf{0.6609} \\
\bottomrule
\end{tabular}
\caption{Performance comparison including gold standard references and different personalization models. \textbf{E\&T}: Engineering \& Tools, \textbf{S\&T}: Science \& Theory, \textbf{L\&S}: Lifestyle \& Society, \textbf{L\&F}: Leisure \& Fandom.}
\label{tab:baselines}
\end{table}

\subsection{Human Validation of CoPA Scoring}
To assess whether CoPA scoring is reasonably aligned with human judgments, we recruited two graduate annotators to independently evaluate 50 randomly sampled question-response pairs using the same 3-point Likert rubric and six-factor criteria as our LLM evaluator. Detailed instructions are provided in Appendix~\ref{sec:Instructions for Human Annotators}. As shown in Table~\ref{tab:human_validation}, the annotators achieved an overall weighted Cohen's Kappa of 0.643, indicating substantial inter-rater agreement. Moreover, the average human ratings showed a Spearman correlation of 0.754 with the LLM scores, suggesting that the proposed scoring scheme is reasonably consistent with human judgments.

\begin{table}[t]
\centering
\small
\renewcommand{\arraystretch}{1.2}
\setlength{\tabcolsep}{4pt}
\begin{tabular}{lcc}
\toprule
\textbf{Factor} & \textbf{Human-Human ($\kappa$)} & \textbf{Human-LLM ($\rho$)} \\
\midrule
CT & 0.622 & 0.733 \\
SA & 0.687 & 0.830 \\
SC & 0.609 & 0.744 \\
CLM & 0.641 & 0.744 \\
MS & 0.663 & 0.683 \\
AMR & 0.637 & 0.788 \\
Overall & 0.643 & 0.754 \\
\bottomrule
\end{tabular}
\caption{Human-Human reports the agreement between the two annotators measured by Cohen's weighted Kappa, and Human-LLM reports the Spearman correlation between the average human ratings and the LLM scores.}
\label{tab:human_validation}
\end{table}

\subsection{Factor Correlation Analysis}
\begin{figure}[h!]
\centering
    \includegraphics[width=0.9\columnwidth]{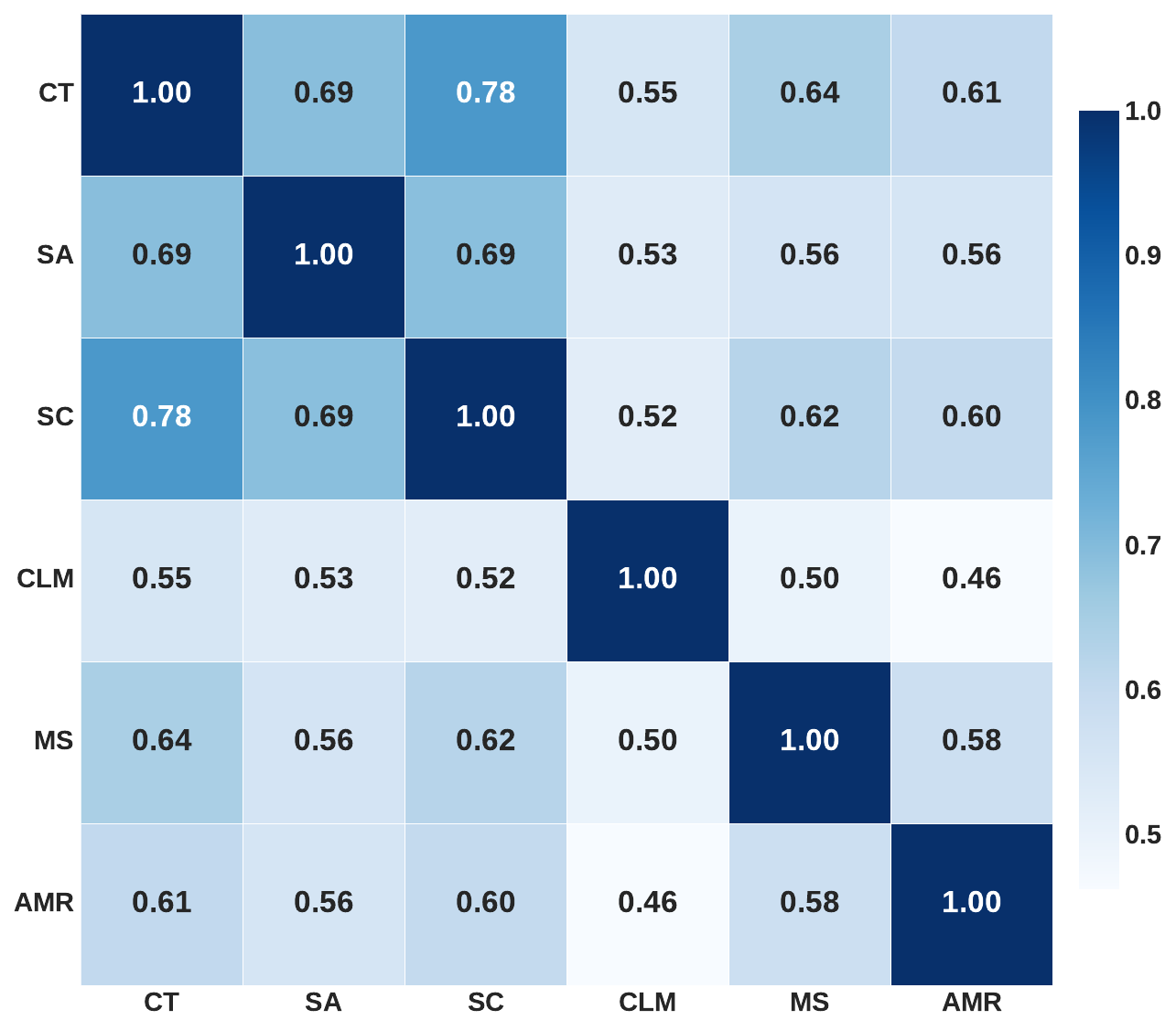}
    \caption{Matrix of Spearman correlations between the factors.}
    \label{fig:correlation_matrix}
\end{figure}

To evaluate the relevance of our factors, We analyzed the inter-factor relationships using the Spearman correlation coefficient \cite{spearman1961proof}, as shown in Figure \ref{fig:correlation_matrix}. The results revealed moderate positive correlations across most factors, indicating that while they collectively contribute to personalization, they capture distinct information without significant redundancy.
Specifically, the peak correlation between Cognitive Trust (CT) and Schema Consistency (SC) highlights a tight dependency: trust relies heavily on alignment with user schemas. However, the coefficient remains below the multicollinearity threshold ($r > 0.9$), confirming their \textit{discriminant validity}—SC represents the establishment \textit{pathway}, while CT reflects the resultant \textit{state}. Conversely, Cognitive Load Management (CLM) displayed the lowest average correlation, underscoring its statistical independence. This confirms that high trust or resonance does not inherently imply low cognitive load. Thus, CLM captures a unique dimension of ``information processing cost,'' validating its necessity as a distinct metric in the \ourM{} benchmark. 

\section{Related Work}

\textbf{Personalized Large Language Models.} Personalization is critical in search \citep{baek2024knowledge,sharma2024generative}, recommendation \cite{wu2024coral,liu2025uqabench}, and text generation \citep{li2024learning,liu2025one}. Existing personalized LLMs fall into two paradigms. The first, non-parametric context augmentation, leverages In-Context Learning to inject user history as external knowledge. Techniques like Retrieval-Augmented Generation \cite{salemi2024optimization,zhuang2024hydra} retrieve relevant behavioral fragments to implicitly mimic user style, while others distill interaction data into explicit User Profiles \cite{chen-etal-2025-deeper,su-etal-2025-personalized} to guide generation. The second paradigm, parametric modeling, internalizes personalized data. While traditional fine-tuning \citep{liu-etal-2025-llms,tan-etal-2024-democratizing} is effective, it suffers from high storage costs and catastrophic forgetting. Consequently, recent lightweight strategies \citep{zhang-etal-2025-personalized,hecontext} propose training-free, dynamic construction of user-specific style matrices at inference time to address these bottlenecks.\\
\textbf{Evaluation of Personalized Generation.} Evaluating alignment with user preferences involves three main paradigms. First, traditional n-gram metrics like BLEU and ROUGE \citep{salemi2024lamp,kumar2024longlamp} measure lexical overlap but struggle with the open-ended, semantic nature of personalized contexts. Second, the mainstream LLM-as-a-Judge approach \citep{salemi-zamani-2025-lamp,dong-etal-2024-llm} utilizes LLMs to assess semantic dimensions (e.g., helpfulness) in a reference-free setting. Third, rule-based heuristics \citep{su-etal-2025-personalized,dai-etal-2024-mpcoder} employ customized measurement standards. However, current metric often lack empirical grounding and fail to elucidate the cognitive determinants driving personalized decisions, highlighting the need for data-driven quantitative standards.
\section{Conclusion}
This paper investigates the Community-Individual Preference Divergence (CIPD) phenomenon within the StackExchange dataset. Through fine-grained user attribution analysis on extensive interaction data, we distilled six core factors. Leveraging these factors, we introduce CoPA, a benchmark designed to evaluate Large Language Models in personalized question answering. Systematic experiments with established baselines demonstrated that our factor-based metrics effectively quantify personalization. Furthermore, empirical results showed that incorporating user profiles significantly enhanced both response quality and preference alignment.

\section*{Acknowledgments}
This work was supported by Zhongguancun Academy (Grant Nos. C20250513 and XTS0025) and the National Natural Science Foundation of China (Grant Nos. 62277017 and 62137001).

\section*{Limitations}
This study has several limitations.\\
First, our work distills six core factors based on the StackExchange dataset. A potential limitation is that these factors may exhibit interdependencies, and their complete orthogonality cannot be guaranteed. Furthermore, the reliance on a single data source might restrict the comprehensiveness of the identified factors. Given the dataset's strong pedagogical and explanatory orientation, the distilled factors may over-represent explanatory preferences prevalent in learning-centric communities. This, in turn, could limit the generalizability of our framework to other personalization scenarios, such as in conversational agents or recommender systems. Future work should therefore aim to integrate heterogeneous data sources to enhance the completeness of factor discovery and to evaluate and adapt our framework across a broader spectrum of application domains.\\
Second, our framework heavily relies on Large Language Models (LLMs) at multiple stages, including principle induction, factor distillation, and evaluation. Although we employed strategies such as context truncation and controlled prompting to ensure the quality of the generated outputs, this dependency may introduce epistemic circularity and inherent model biases. A crucial direction for improving the framework's robustness is to incorporate a human-in-the-loop validation mechanism.\\
Third, the stability and reliability of the current "LLM-as-a-Judge" mechanism are susceptible to the intrinsic biases of LLMs. Consequently, future research will prioritize two avenues: 1) developing more sophisticated and fine-grained Judge Models, and 2) operationalizing the abstract theoretical factors into concrete, measurable metrics under a more rigorous theoretical grounding to enhance the objectivity and accuracy of the evaluation.
\section*{Ethics Statement}
Potential ethical concerns arise from the use of datasets collected from open-source web platforms, which may carry the risk of privacy leakage. To mitigate this, we have manually sanitized the datasets by redacting all Personally Identifiable Information, such as usernames and identifiers. Furthermore, we acknowledge that content generated by Large Language Models (LLMs) is susceptible to hallucinations and ethical pitfalls. Consequently, a rigorous evaluation of the reliability and safety of the model outputs is essential.

\bibliography{reference}
\clearpage
\appendix
\setcounter{table}{0}
\setcounter{figure}{0}
\renewcommand{\thetable}{A\arabic{table}}
\renewcommand{\thefigure}{A\arabic{figure}}
\newpage
\section{Question Distribution}\label{sec:Question Distribution}
\begin{figure*}[p!]
\centering
    \includegraphics[width=2.0\columnwidth]{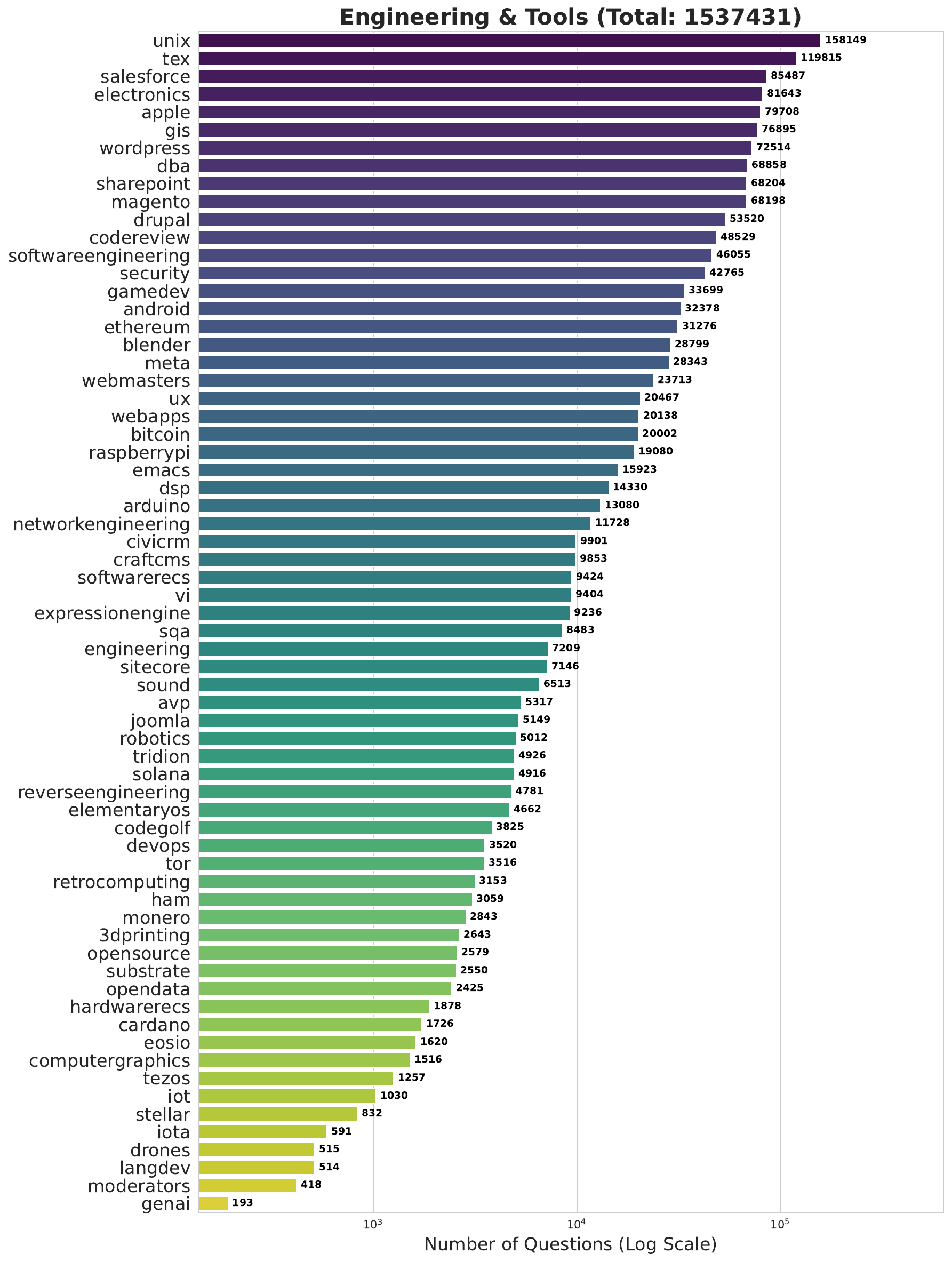}
    \caption{The distribution of questions across various domains in the Engineering\&Tools.}
    \label{fig:domain_question_count_Engineering_and_Tools}
\end{figure*}

\begin{figure*}[p]
\centering
    \includegraphics[width=2.0\columnwidth]{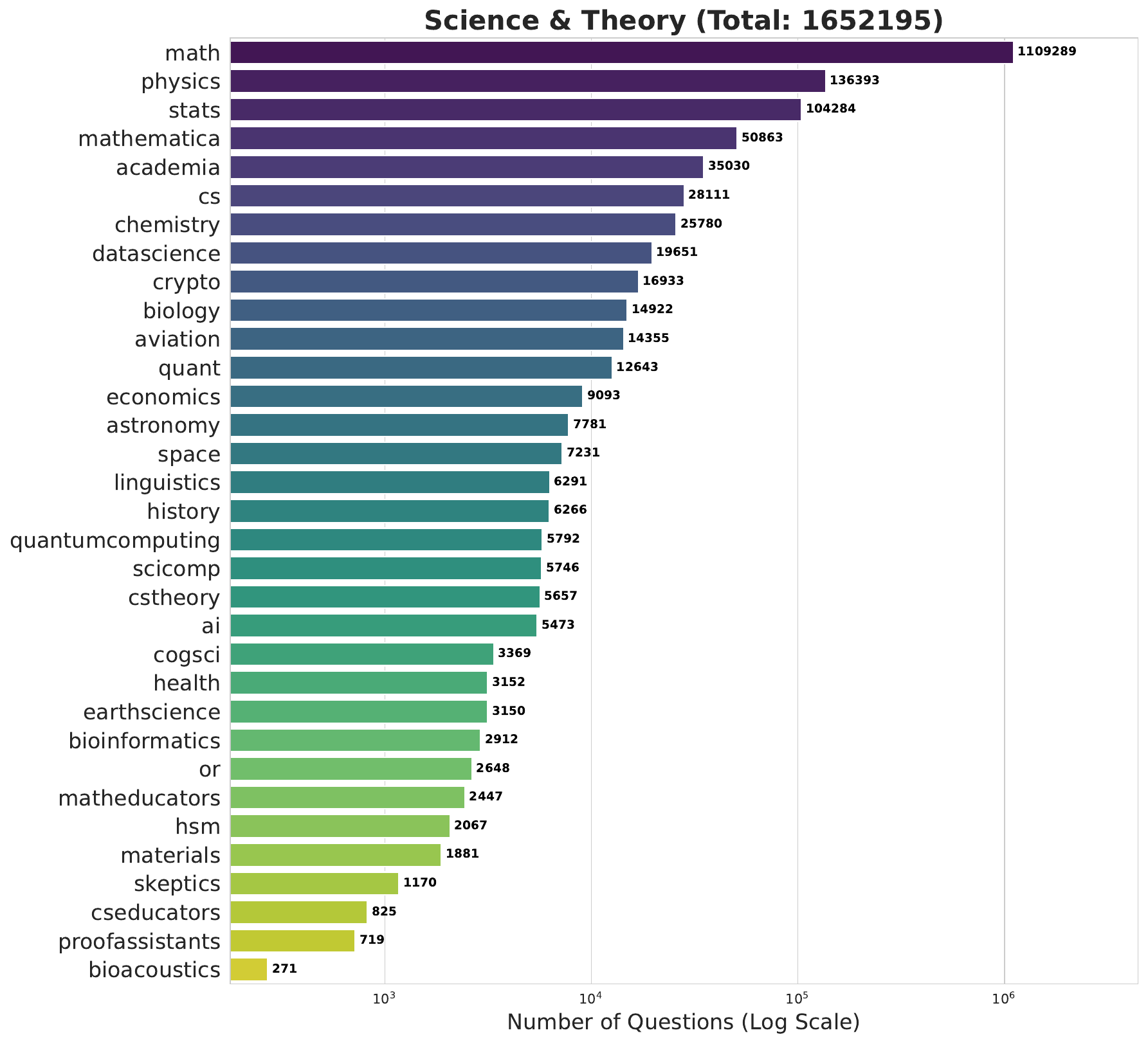}
    \caption{The distribution of questions across various domains in the Science\&Theory.}
    \label{fig:domain_question_count_Science_and_Theory}
\end{figure*}

\begin{figure*}[p]
\centering
    \includegraphics[width=2.0\columnwidth]{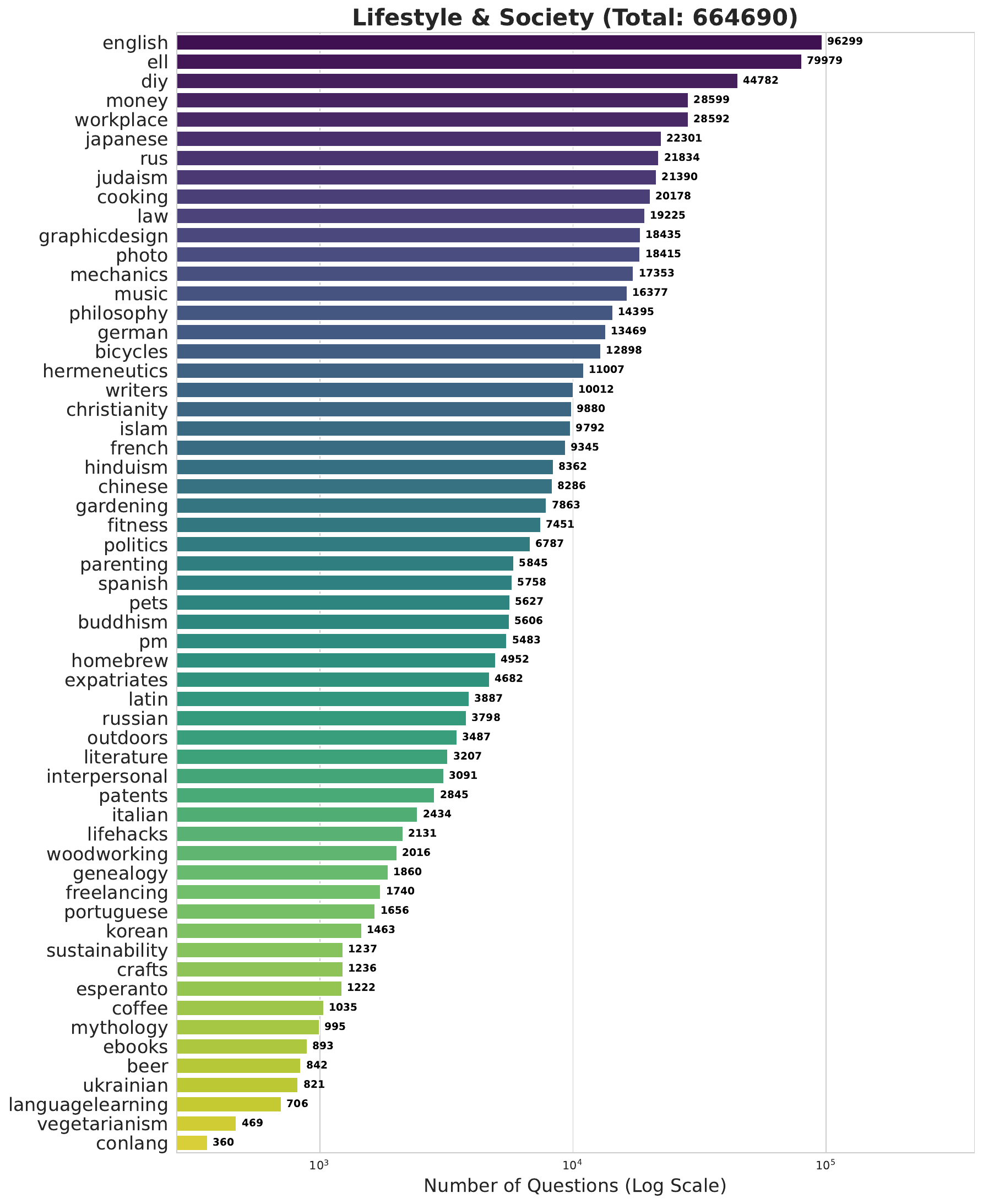}
    \caption{The distribution of questions across various domains in the Lifestyle\&Society.}
    \label{fig:domain_question_count_Lifestyle_and_Society}
\end{figure*}

\begin{figure*}[p]
\centering
    \includegraphics[width=2.0\columnwidth]{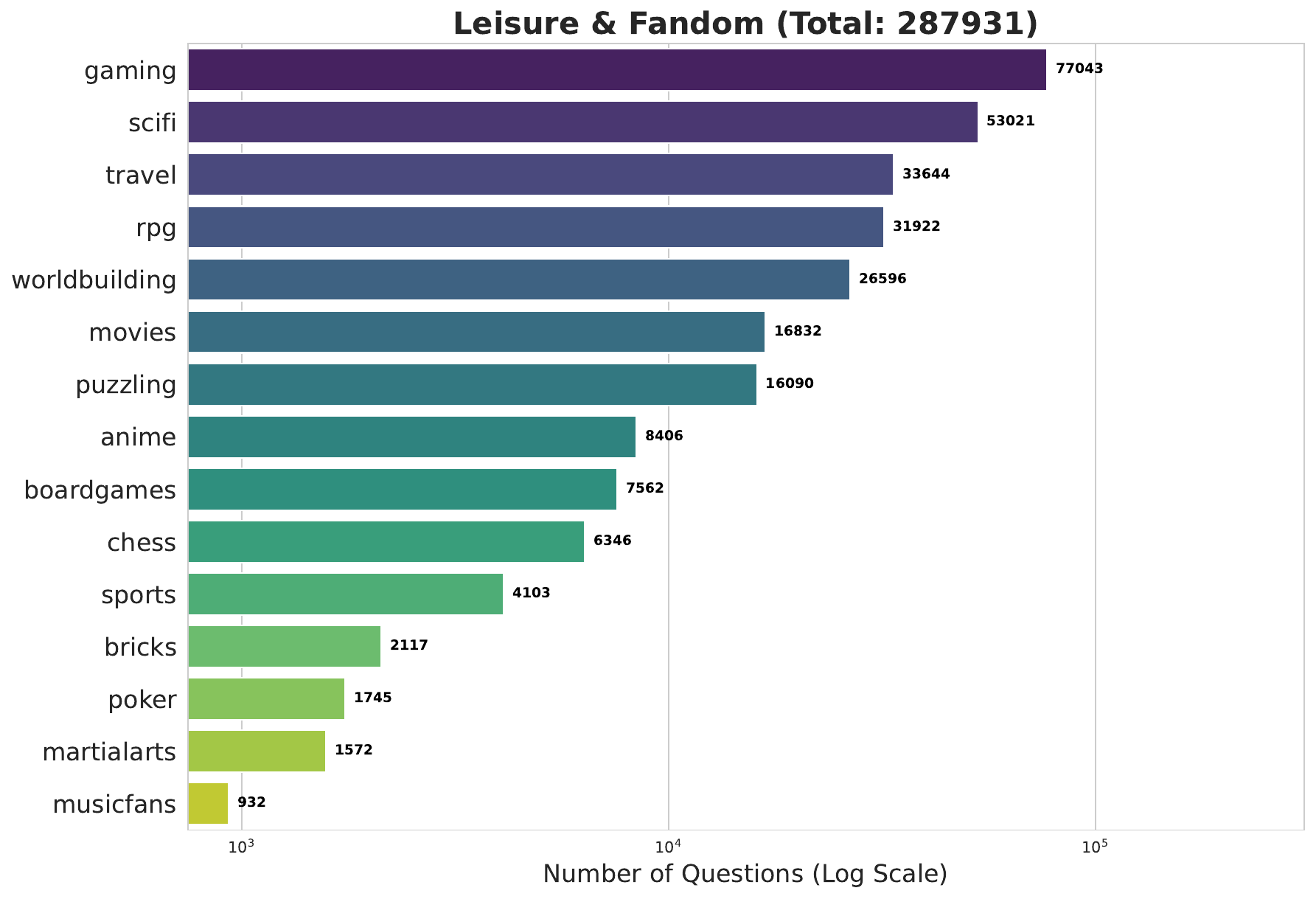}
    \caption{The distribution of questions across various domains in the Leisure\&Fandom.}
    \label{fig:domain_question_count_Leisure_and_Fandom}
\end{figure*}

Figure \ref{fig:domain_question_count_Engineering_and_Tools} shows the distribution of questions across various domains in the Engineering\&Tools category. Figure \ref{fig:domain_question_count_Science_and_Theory} shows the distribution of questions across various domains in the Science\&Theory category. Figure \ref{fig:domain_question_count_Lifestyle_and_Society} shows the distribution of questions across various domains in the Lifestyle\&Society category. Figure \ref{fig:domain_question_count_Leisure_and_Fandom} shows the distribution of questions across various domains in the Leisure\&Fandom category.

\section{Factor Mining}\label{sec:Factor Mining Prompt}
Figure \ref{fig:Extract rationale Prompt} shows the prompt used for Rationale Extraction. Figure \ref{fig:Factor Mine Prompt} shows the prompt used for Factor Mining.

\begin{figure}[p]
    \small
    \centering
    \begin{tcolorbox}[colframe=RoyalBlue, colback=white]
    \textbf{System:}
    \\
    You are an experienced educationalist who has researched in the field of education for many years. Your specialty is to analyze a user's psychology and choices from a pedagogical perspective to understand their learning behavior and needs.
    \\
    \vspace{1mm} 
    
    \textbf{User:}
    \\
    User's historical question:
    \\
    \{Historical\_questions\}
    \\
    User's current question:
    \\
    \{Current\_question\}
    \\
    The answer to the current question:
    \\
    \{Answers\}
    \\
    The 'IsAccept' field in the answer data signifies user acceptance. Your task is to analyze the reasons why the user accepted this specific answer over other alternatives that had higher scores.
    \\
    \vspace{1mm}
    
    The response should be in JSON format:
    \\
    \{\\
      \hspace*{1em}"reasons": [\\
        \hspace*{2em}\{\\
          \hspace*{3em}"reason": "",\\
          \hspace*{3em}"theory":"",\\
          \hspace*{3em}"explanation": ""\\
        \hspace*{2em}\},\\
        \hspace*{2em}\{\\
          \hspace*{3em}"reason": "",\\
          \hspace*{3em}"theory": "",\\
          \hspace*{3em}"explanation": ""\\
        \hspace*{2em}\}\\
      \hspace*{1em}],\\
      \hspace*{1em}"global\_explanation": ""\\
    \}
    \\
    Requirements:
    \\
    1. Each reason should be concise and not exceed five words.
    \\
    2. Each reason should be grounded in an educational or psychological framework, incorporating appropriate academic terminology.
    \\
    3. There is no limit to the number of reasons.
    \\
    4. The theory embodies the educational principles reflected in this reason.
    \\
    5. The 'global\_explanation' should provide an overall summary of the reasons listed above.
    \end{tcolorbox}
    \caption{Extract rationale Prompt.}
    \label{fig:Extract rationale Prompt}
\end{figure}

\begin{figure}[h!]
    \small
    \centering
    \begin{tcolorbox}[colframe=RoyalBlue, colback=white]
    \textbf{System:}
    \\
    You are a researcher in the field of education with expertise in pedagogy, psychology, and linguistics.
    \\
    \vspace{1mm}
    
    \textbf{User:}
    \\
    The pool of reasons for users' choices.
    \\
    \{Reasons\}
    \\
    The pool of reasons contains the rationales for why different users selected their answers. Your task is to analyze these rationales from an educational perspective to derive a set of quantifiable factors.
    \\
    \vspace{1mm}
    
    The response should be in JSON format:
    \\
    \{\\
      \hspace*{1em}"factors": [\\
        \hspace*{2em}\{\\
          \hspace*{3em}"factor": "",\\
          \hspace*{3em}"example": "",\\
          \hspace*{3em}"explanation": ""\\
        \hspace*{2em}\},\\
        \hspace*{2em}\{\\
          \hspace*{3em}"factor": "",\\
          \hspace*{3em}"example": "",\\
          \hspace*{3em}"explanation": ""\\
        \hspace*{2em}\}\\
      \hspace*{1em}],\\
      \hspace*{1em}"global\_explanation": ""\\
    \}
    \\
    Requirements:
    \\
    1. You can use academic terminology to describe these factors.
    \\
    2. The example section illustrates the conflicting reasons you found. It must contain at least five pairs of contrasting examples, with one supporting a choice (a positive example) and one opposing it (a negative example).
    \\
    3. There is no limit to the number of factors.
    \\
    4. The 'global\_explanation' should provide an overall summary of the factor listed above.
    \end{tcolorbox}
    \caption{Factor Mine Prompt.}
    \label{fig:Factor Mine Prompt}
\end{figure}

\subsection{Manual Inspection Criteria for Reason Analysis}\label{sec:Manual Inspection Criteria for Reason Analysis}
We evaluate the Rationale Extraction based on the following criteria:
\begin{description}
    \item[1 - Pass] if the rationale is both logically coherent and theoretically sound. This requires the model to accurately identify an underlying motive and cite a relevant theory (e.g., ``Social Identity Theory,'' ``Cognitive Dissonance'') that strongly supports it. The output must contain no factual errors.

    \item[0 - Fail] if the rationale exhibits any of the following issues:
    \begin{enumerate}[label=\arabic*., nosep]
        \item \textbf{Hallucination or Contradiction:} The inferred motive is irrelevant to, or contradicts, the user's provided conversation history.
        \item \textbf{Theoretical Error:} The model fabricates a non-existent theory or concept.
        \item \textbf{Inappropriate Application of Theory:} A valid theory is cited, but it is completely mismatched with the inferred motive (e.g., applying ``Economic Utility Theory'' to explain a rationale for ``emotional counseling'').
    \end{enumerate}
\end{description}

\subsection{Factors and Their Associated Causal Cases}\label{sec:Factors and Their Associated Causal Cases}
Table\ref{tab:case1}, \ref{tab:case2}, and \ref{tab:case3} show three examples of factors associated with the reasons.

We evaluate the factors based on the following criteria:
\begin{description}
    \item[1.] Whether the factors are subject to cognitive biases.
    \item[2.] Whether these factors constitute dimensions that influence an individual's decision-making process.
    \item[3.] Whether the factors are generalizable across different populations, cultures, and contexts.
\end{description}
\begin{table*}[h!]
\centering
\caption{Case1}
\label{tab:case1}

\begin{tabularx}{\textwidth}{@{} >{\raggedright}p{3.2cm} >{\raggedright}p{2.8cm} >{\raggedright\arraybackslash}X @{}}
\toprule
\textbf{Reason} & \textbf{Theory} & \textbf{Explanation} \\
\midrule
Narrative gap resolution & Cognitive closure & The answer filled a specific gap left by the anime adaptation, providing the user with a clear explanation from the manga, thus satisfying their need for a complete mental model of the story. \\
\addlinespace 

Source-authoritative detail & Epistemic trust & The response referenced information from the primary source material (manga), which the user perceives as more authoritative and trustworthy, increasing acceptance likelihood. \\
\addlinespace

Canon consistency & Schema theory & The explanation aligned with the user's existing narrative schema of the series, integrating seamlessly without contradictions, which aids in knowledge assimilation. \\
\addlinespace

Context-specific relevance & Situated learning & The answer directly addressed the specific context of the user's question (Touka's ability in Inari Konkon), creating immediate perceived practical relevance. \\
\midrule
\textbf{Global Explanation} & \multicolumn{2}{p{\dimexpr\linewidth-3.2cm-2\tabcolsep\relax}}{The user accepted this answer because it directly addressed a specific narrative gap from the anime by referencing canonical manga evidence, which satisfied their cognitive need for closure. The information was consistent with their existing knowledge schema, delivered from what they perceived as an authoritative source, and situated within the precise narrative context they were curious about, all of which increased trust, relevance, and integration into their understanding.} \\
\midrule

\textbf{Factors} & \multicolumn{2}{p{\dimexpr\linewidth-3.2cm-2\tabcolsep\relax}}{Cognitive Trust, Situational Anchoring, Schema Consistency} \\
\bottomrule
\end{tabularx}
\end{table*}

\begin{table*}[h!]
\centering
\caption{Case2}
\label{tab:case2}

\begin{tabularx}{\textwidth}{@{} >{\raggedright}p{3.2cm} >{\raggedright}p{2.8cm} >{\raggedright\arraybackslash}X @{}}
\toprule
\textbf{Reason} & \textbf{Theory} & \textbf{Explanation} \\
\midrule
Clear conceptual differentiation & Cognitive Load Theory & The answer helped reduce intrinsic cognitive load by clearly distinguishing EFHW and long wire antennas, which addressed the user's confusion and allowed them to process the technical differences more efficiently. \\
\addlinespace

Direct application insight & Situated Learning Theory & The answer connected principles to the user's real-world scenario (hotel room limitations), enabling context-based understanding that made the information more relevant and actionable. \\
\addlinespace

Mechanism-focused explanation & Constructivist Learning Theory & The answer explained how multiple loops would affect inductance, SWR curve, and radiation pattern, enabling the user to build mental models from prior knowledge. \\
\addlinespace

Expectation management & Metacognition & By acknowledging the inevitable compromise in performance, the answer guided the user's realistic assessment of trade-offs, improving self-regulation in decision-making. \\
\midrule
\textbf{Global Explanation} & \multicolumn{2}{p{\dimexpr\linewidth-3.2cm-2\tabcolsep\relax}}{The user accepted the answer because it reduced conceptual confusion between antenna types, contextualized the explanation to their space-limited operating environment, provided clear technical mechanisms, and managed expectations of performance. From a pedagogical perspective, the answer optimized cognitive load, fostered situated learning, supported constructivist model-building, and enhanced metacognitive awareness—leading to a response that was both technically informative and psychologically reassuring.} \\
\midrule

\textbf{Factors} & \multicolumn{2}{p{\dimexpr\linewidth-3.2cm-2\tabcolsep\relax}}{Cognitive Load Management, Situational Anchoring, Schema Consistency, Metacognitive Scaffolding} \\
\bottomrule
\end{tabularx}
\end{table*}

\begin{table*}[h!]
\centering
\caption{Case3}
\label{tab:case3}
\begin{tabularx}{\textwidth}{@{} >{\raggedright}p{3.2cm} >{\raggedright}p{2.8cm} >{\raggedright\arraybackslash}X @{}}
\toprule
\textbf{Reason} & \textbf{Theory} & \textbf{Explanation} \\
\midrule
Growth mindset emphasis & Carol Dweck's Growth Mindset Theory & The accepted answer likely stressed that intelligence and skill can develop with effort, aligning with Dweck's research which shows children persist more when they view challenges as opportunities to grow rather than as fixed measures of ability. \\
\addlinespace
Intrinsic motivation cultivation & Self-Determination Theory & By focusing on the satisfaction of mastering a difficult task, the answer may have encouraged autonomy and competence, supporting self-determination theory’s emphasis on internal rewards over external pressures. \\
\addlinespace
Scaffolded learning approach & Vygotsky's Zone of Proximal Development & The answer may have suggested breaking tasks into achievable steps, enabling the child to succeed incrementally while receiving guidance, which aligns with Vygotsky’s principle of supporting learners just beyond their current capability. \\
\addlinespace
Positive reinforcement use & Skinner's Operant Conditioning & Likely included recommendations of rewarding effort and persistence, teaching the child to associate trying again with positive outcomes, reinforcing the desired behavior. \\
\midrule
\textbf{Global Explanation} & \multicolumn{2}{p{\dimexpr\linewidth-3.2cm-2\tabcolsep\relax}}{The user accepted the answer because it acknowledged the psychological roots of giving up after failure and addressed them through proven educational frameworks. By combining growth mindset principles, intrinsic motivation development, scaffolded support, and positive reinforcement, the answer offered a multi-dimensional strategy that resonates with both pedagogical and developmental psychology approaches, ensuring practical guidance aligned with the child's stage of cognitive and emotional growth.} \\
\midrule
\textbf{Factors} & \multicolumn{2}{p{\dimexpr\linewidth-3.2cm-2\tabcolsep\relax}}{Metacognitive Scaffolding, Affective and Motivational Resonance, Schema Consistency} \\
\bottomrule
\end{tabularx}
\end{table*}

\section{Factor Label}\label{sec:Factor Label Prompt}
Figure \ref{fig:Factor Label Prompt} shows the prompt used for factor label. Figure \ref{fig:Evaluation Prompt system} shows the prompt used for evaluation .
\begin{figure}[h!]
    \small
    \centering
    \begin{tcolorbox}[colframe=RoyalBlue, colback=white]
    \textbf{System:}
    \\
    Role: You are an interdisciplinary domain expert.
    Your task is to evaluate the specific emphasis the user places on the following dimensions regarding the current question, based on an analysis of the user’s historical query patterns and the response to the immediate question.
    \\
    \vspace{1mm}
    
    Evaluation Dimensions:
    \begin{enumerate}[leftmargin=*, nosep]
        \item Cognitive Trust: What are the user's epistemic requirements regarding the credibility, reliability, and verifiability of the information?
        \item Situational Anchoring: To what extent does the user require the response to be contextually aligned, practically applicable, or specific to a given scenario?
        \item Schema Consistency: What is the nature of the user's existing prior knowledge and mental models (and how should the new information align with them)?
        \item Cognitive Load Management: What are the user's constraints regarding information processing capacity and their tolerance for complexity?
        \item Metacognitive Scaffolding: What are the user's requirements for structural guidance to facilitate higher-order understanding and self-regulated learning?
        \item Affective and Motivational Resonance: What are the user's expectations regarding emotional engagement and motivational alignment within the response?
    \end{enumerate}
    \vspace{1mm}
    
    \textbf{User:}
    \\
    User's Historical Questions:
    \\
    \{Historical\_questions\}
    \\
    User's Current Question:
    \\
    \{Current\_question\}
    \\
    The answer to the current question:
    \\
    \{Answer\}
    \\
    \vspace{1mm}
    
    The response should be in JSON format:
    \\
    \{\\
        \hspace*{1em}"Cognitive Trust": \{\\
            \hspace*{2em}"description": "", "explanation": ""\\
        \hspace*{1em}\},\\
        \hspace*{1em}"Situational Anchoring": \{\\
            \hspace*{2em}"description": "", "explanation": ""\\
        \hspace*{1em}\},\\
        \hspace*{1em}"Schema Consistency": \{\\
            \hspace*{2em}"description": "", "explanation": ""\\
        \hspace*{1em}\},\\
        \hspace*{1em}"Cognitive Load Management": \{\\
            \hspace*{2em}"description": "", "explanation": ""\\
        \hspace*{1em}\},\\
        \hspace*{1em}"Metacognitive Scaffolding": \{\\
            \hspace*{2em}"description": "", "explanation": ""\\
        \hspace*{1em}\},\\
        \hspace*{1em}"Affective and Motivational Resonance": \{\\
            \hspace*{2em}"description": "", "explanation": ""\\
        \hspace*{1em}\}\\
    \}
    \\
    Requirements:
    \\
    1. Each description must be tailored to the user's specific circumstances.
    \\
    2. Each description should provide a specific and accurate summary of the user's personal profile, with no word limit.
    \\
    3. The explanation serves as the rationale for the description provided above.
    \\
    4. Ensure the output adheres to the specified format.
    \end{tcolorbox}
    \caption{Factor Label Prompt.}
    \label{fig:Factor Label Prompt}
\end{figure}
\newpage
\begin{figure*}[t]
    \small
    \centering
    \begin{tcolorbox}[colframe=RoyalBlue, colback=white]
    \textbf{System:}
    \\
    Role: You are a fair and insightful judge with exceptional reasoning and analytical abilities. 
    Your task is to evaluate whether the response to the user's question aligns with the user's factor profile.
    \\
    \vspace{1mm}
    
    \textbf{Evaluation Criteria (The 6 Dimensions):}
    \begin{enumerate}[leftmargin=*, nosep]
        \item Cognitive Trust: What are the user's epistemic requirements regarding the credibility, reliability, and verifiability of the information?
        \item Situational Anchoring: To what extent does the user require the response to be contextually aligned, practically applicable, or specific to a given scenario?
        \item Schema Consistency: What is the nature of the user's existing prior knowledge and mental models (and how should the new information align with them)?
        \item Cognitive Load Management: What are the user's constraints regarding information processing capacity and their tolerance for complexity?
        \item Metacognitive Scaffolding: What are the user's requirements for structural guidance to facilitate higher-order understanding and self-regulated learning?
        \item Affective and Motivational Resonance: What are the user's expectations regarding emotional engagement and motivational alignment within the response?
    \end{enumerate}
    \vspace{1mm}
    
    \textbf{Scoring Rubric (3-point Likert scale):}
    \begin{itemize}[leftmargin=*]
        \item 0 (Mismatch): The response actively violates the user's preference or completely ignores a high-priority requirement defined in the profile.
        \item 1 (Partial Match): The response addresses the requirement but lacks depth, or only partially aligns with the user's preference.
        \item 2 (Full Match): The response perfectly adapts to the user's constraints and preferences described in the profile.
    \end{itemize}

    \textbf{User:}
    \\
    \textbf{Input Data:}
    \\
    <user\_factor\_profile>
    \\
    \{factors\_profile\}
    \\
    </user\_factor\_profile>
    \\
    \vspace{1mm}
    
    <question>
    \\
    \{question\}
    \\
    </question>
    \\
    \vspace{1mm}
    
    <response\_to\_evaluate>
    \\
    \{response\}
    \\
    </response\_to\_evaluate>
    \\
    \vspace{1mm}
    
    The response should be in JSON format:
    \\
    \{\\
        \hspace*{1em}"Cognitive Trust": \{
            "score": 0,
            "reasoning": "Brief explanation..."
        \},\\
        \hspace*{1em}"Situational Anchoring": \{
            "score": 0,
            "reasoning": "Brief explanation..."
        \},\\
        \hspace*{1em}"Schema Consistency": \{
            "score": 0,
            "reasoning": "Brief explanation..."
        \},\\
        \hspace*{1em}"Cognitive Load Management": \{
            "score": 0,
            "reasoning": "Brief explanation..."
        \},\\
        \hspace*{1em}"Metacognitive Scaffolding": \{
            "score": 0,
            "reasoning": "Brief explanation..."
        \},\\
        \hspace*{1em}"Affective and Motivational Resonance": \{
            "score": 0,
            "reasoning": "Brief explanation..."
        \}\\
    \}
    \\
    \vspace{1mm}
    
    Requirements:
    \begin{enumerate}[leftmargin=*, nosep]
        \item Analyze the match between the <user\_factor\_profile> and <response\_to\_evaluate> for EACH factor.
        \item Assign a score of 0, 1, or 2. \textbf{Important: The 'score' field must be a raw INTEGER type (int), do not use strings.} (e.g., output 1, not "1").
        \item Provide a brief reasoning for your score.
        \item Output the result in strict JSON format.
    \end{enumerate}
    \end{tcolorbox}
    \caption{Evaluation Prompt.}
    \label{fig:Evaluation Prompt system}
\end{figure*}

\section{Effectiveness of Factors}\label{sec:Effectiveness of Factors}
Figures \ref{fig:Direct Evaluation Prompt}, \ref{fig:CoT Evaluation Prompt}, and \ref{fig:Random Evaluation Prompt} show the prompts used for the Direct, CoT, and Random baselines, respectively. Table~\ref{tab:Comparison of Evaluation Methodologies_appendix} reports the main experimental results when the factor profile for evaluation is constructed from the current CIPD question.

\begin{figure*}[t]
    \small
    \centering
    \begin{tcolorbox}[colframe=RoyalBlue, colback=white]
    \textbf{System:}
    \\
    Role: You are a fair and insightful judge with exceptional reasoning and analytical abilities.
    Your task is to evaluate whether the response to the user's question aligns with the user's profile.
    \\
    \vspace{1mm}

    \textbf{Scoring Rubric (3-point Likert scale):}
    \begin{itemize}[leftmargin=*]
        \item 0 (Mismatch): The response actively violates the user's preference or completely ignores a high-priority requirement defined in the profile.
        \item 1 (Partial Match): The response addresses the requirement but lacks depth, or only partially aligns with the user's preference.
        \item 2 (Full Match): The response perfectly adapts to the user's constraints and preferences described in the profile.
    \end{itemize}
    \vspace{1mm}

    \textbf{User:}
    \\
    \textbf{Input Data:}
    \\
    <user\_profile>
    \\
    \{user\_profile\}
    \\
    </user\_profile>
    \\
    \vspace{1mm}

    <question>
    \\
    \{question\}
    \\
    </question>
    \\
    \vspace{1mm}

    <response\_to\_evaluate>
    \\
    \{response\}
    \\
    </response\_to\_evaluate>
    \\
    \vspace{1mm}

    The response should be in JSON format:
    \\
    \{\\
    \hspace*{1em}"score": 0,\\
    \hspace*{1em}"reasoning": ""\\
    \}
    \\
    \vspace{1mm}

    Requirements:
    \begin{enumerate}[leftmargin=*, nosep]
        \item Analyze the match between the <user\_profile> and <response\_to\_evaluate>.
        \item Assign a score of 0, 1, or 2. \textbf{Important: The `score` field must be a raw INTEGER type (int), do not use strings.}
        \item Provide a brief reasoning for your score.
        \item Output the result in strict JSON format.
    \end{enumerate}
    \end{tcolorbox}
    \caption{Direct Evaluation Prompt.}
    \label{fig:Direct Evaluation Prompt}
\end{figure*}

\begin{figure*}[p]
    \scriptsize
    \centering
    \begin{tcolorbox}[colframe=RoyalBlue, colback=white]
    \textbf{System:}
    \\
    Role: You are a fair and insightful judge with exceptional reasoning and analytical abilities.
    Your task is to evaluate whether the response to the user's question aligns with the user's profile.
    \\
    \vspace{1mm}

    \textbf{Scoring Rubric (3-point Likert scale):}
    \begin{itemize}[leftmargin=*]
        \item 0 (Mismatch): The response actively violates the user's preference or completely ignores a high-priority requirement defined in the profile.
        \item 1 (Partial Match): The response addresses the requirement but lacks depth, or only partially aligns with the user's preference.
        \item 2 (Full Match): The response perfectly adapts to the user's constraints and preferences described in the profile.
    \end{itemize}

    \textbf{Evaluation Instructions (Chain-of-Thought):}
    \begin{enumerate}[leftmargin=*, nosep]
        \item Step 1 -- Profile Analysis: Identify the key preferences and requirements revealed by the user's historical questions.
        \item Step 2 -- Response Analysis: Examine what the response actually provides (e.g., depth, style, format, domain focus).
        \item Step 3 -- Alignment Assessment: Compare the user's profile against the response. Note specific matches or mismatches.
        \item Step 4 -- Score Decision: Based on the above analysis, assign a score of 0, 1, or 2.
    \end{enumerate}
    \vspace{1mm}

    \textbf{Output Format:}
    \\
    Return a single JSON object with the following fields:
    \\
    \{\\
    \hspace*{1em}"chain\_of\_thought": "Your step-by-step reasoning following Steps 1--4",\\
    \hspace*{1em}"score": 0,\\
    \hspace*{1em}"reasoning": "One-sentence summary of the final judgement"\\
    \}
    \\
    IMPORTANT: The `score` field must be a raw INTEGER type (int), not a string.
    \\
    \vspace{1mm}

    \textbf{Demonstration Examples:}
    \begin{enumerate}[leftmargin=*, nosep]
        \item \textbf{Example 1 -- Score: 0 (Mismatch).}
        \\
        <user\_profile>
        1. How does backpropagation work mathematically in deep neural networks?
        2. What is the derivation of the attention mechanism in Transformers?
        3. Can you explain the proof of the universal approximation theorem?
        4. What are the theoretical guarantees of gradient descent convergence?
        5. How is the VC dimension defined and why does it bound generalization error?
        </user\_profile>
        \\
        <question>What is a neural network?</question>
        \\
        <response\_to\_evaluate>
        A neural network is a machine learning model loosely inspired by the brain. It consists of layers of nodes (neurons) that transform input data step by step. Popular frameworks include TensorFlow and PyTorch.
        </response\_to\_evaluate>
        \\
        Expected output:
        \\
        \{\\
        \hspace*{1em}"chain\_of\_thought": "Step 1 - Profile Analysis: All historical questions demand mathematical derivations and theoretical proofs, indicating an expert user. Step 2 - Response Analysis: The response is a beginner-level intuitive overview with no mathematics. Step 3 - Alignment Assessment: Direct conflict --- the user expects formal rigor; the response delivers none. Step 4 - Score Decision: Clear mismatch.",\\
        \hspace*{1em}"score": 0,\\
        \hspace*{1em}"reasoning": "The response provides a superficial overview that completely ignores the user's demonstrated need for rigorous mathematical depth."\\
        \}

        \item \textbf{Example 2 -- Score: 2 (Full Match).}
        \\
        <user\_profile>
        1. I am new to Python, how do I print hello world?
        2. What does the if-else statement do in Python?
        3. How do I create a list in Python and add items to it?
        4. What is a for loop and how do I use it?
        5. Can you explain what a function is in Python with a simple example?
        </user\_profile>
        \\
        <question>How do I read a file in Python?</question>
        \\
        <response\_to\_evaluate>
        To read a file in Python, use the built-in open() function: with open("myfile.txt", "r") as f: content = f.read() print(content) The with statement closes the file automatically. You can also read line by line using a for loop, which you already know!
        </response\_to\_evaluate>
        \\
        Expected output:
        \\
        \{\\
        \hspace*{1em}"chain\_of\_thought": "Step 1 - Profile Analysis: All questions are beginner-level Python syntax questions; user needs simple language and concrete examples. Step 2 - Response Analysis: Clear code example, plain language, connects to prior knowledge of for loops. Step 3 - Alignment Assessment: Perfect match --- accessible, concrete, no unnecessary complexity. Step 4 - Score Decision: Full match.",\\
        \hspace*{1em}"score": 2,\\
        \hspace*{1em}"reasoning": "The response perfectly adapts to the beginner profile with simple language, a practical example, and a connection to prior knowledge."\\
        \}

        \item \textbf{Example 3 -- Score: 1 (Partial Match).}
        \\
        <user\_profile>
        1. What are the latest advancements in large language model alignment research?
        2. How does RLHF compare to Constitutional AI in terms of safety properties?
        3. What are the known failure modes of reward hacking in AI systems?
        4. Can you explain the debate around instrumental convergence in AI safety?
        5. What is the current state of interpretability research for transformer models?
        </user\_profile>
        \\
        <question>What are the risks of deploying large language models in production?</question>
        \\
        <response\_to\_evaluate>
        Deploying LLMs in production carries risks such as hallucination, bias, high compute costs, and potential misuse. Implement guardrails, monitoring, and human oversight.
        </response\_to\_evaluate>
        \\
        Expected output:
        \\
        \{\\
        \hspace*{1em}"chain\_of\_thought": "Step 1 - Profile Analysis: User specializes in AI safety research --- RLHF, Constitutional AI, reward hacking, instrumental convergence. Step 2 - Response Analysis: Covers generic production risks at a business level; no mention of alignment-specific failure modes. Step 3 - Alignment Assessment: Topically related but misses the researcher-level depth the user expects (e.g., deceptive alignment, distributional shift). Step 4 - Score Decision: Partial match.",\\
        \hspace*{1em}"score": 1,\\
        \hspace*{1em}"reasoning": "The response addresses surface-level risks but lacks the technical alignment-research depth the user's profile requires."\\
        \}
    \end{enumerate}
    \vspace{1mm}

    \textbf{User:}
    \\
    \textbf{Input Data:}
    \\
    <user\_profile>
    \\
    \{user\_profile\}
    \\
    </user\_profile>
    \\
    \vspace{1mm}
    <question>
    \\
    \{question\}
    \\
    </question>
    \\
    \vspace{1mm}
    <response\_to\_evaluate>
    \\
    \{response\}
    \\
    </response\_to\_evaluate>
    \\
    \vspace{1mm}
    Follow the Chain-of-Thought steps above and return your evaluation as a single JSON object with fields: "chain\_of\_thought", "score" (integer 0/1/2), and "reasoning".
    \end{tcolorbox}
    \caption{CoT Evaluation Prompt.}
    \label{fig:CoT Evaluation Prompt}
\end{figure*}

\begin{figure*}[t]
    \small
    \centering
    \begin{tcolorbox}[colframe=RoyalBlue, colback=white]
    \textbf{System:}
    \\
    Role: You are a fair and insightful judge with exceptional reasoning and analytical abilities.
    Your task is to evaluate whether the response to the user's question aligns with the user's factor profile.
    \\
    \vspace{1mm}

    \textbf{Scoring Rubric (3-point Likert scale):}
    \begin{itemize}[leftmargin=*]
        \item 0 (Mismatch): The response actively violates the user's preference or completely ignores a high-priority requirement defined in the profile.
        \item 1 (Partial Match): The response addresses the requirement but lacks depth, or only partially aligns with the user's preference.
        \item 2 (Full Match): The response perfectly adapts to the user's constraints and preferences described in the profile.
    \end{itemize}
    \vspace{1mm}

    \textbf{User:}
    \\
    \textbf{Input Data:}
    \\
    <user\_factor\_profile>
    \\
    \{factors\_profile\}
    \\
    </user\_factor\_profile>
    \\
    \vspace{1mm}

    <question>
    \\
    \{question\}
    \\
    </question>
    \\
    \vspace{1mm}

    <response\_to\_evaluate>
    \\
    \{response\}
    \\
    </response\_to\_evaluate>
    \\
    \vspace{1mm}

    The response should be in JSON format:
    \par
    \begingroup
    \ttfamily
    \texttt{[}\par
    \hspace*{1em}\texttt{\{}\par
    \hspace*{2em}\texttt{"factor": "",}\par
    \hspace*{2em}\texttt{"score": 0,}\par
    \hspace*{2em}\texttt{"reasoning": "Brief explanation..."}\par
    \hspace*{1em}\texttt{\},}\par
    \hspace*{1em}\texttt{\{}\par
    \hspace*{2em}\texttt{"factor": "",}\par
    \hspace*{2em}\texttt{"score": 0,}\par
    \hspace*{2em}\texttt{"reasoning": "Brief explanation..."}\par
    \hspace*{1em}\texttt{\},}\par
    \hspace*{1em}\texttt{...}\par
    \texttt{]}\par
    \endgroup
    \vspace{1mm}

    Requirements:
    \begin{enumerate}[leftmargin=*, nosep]
        \item Each factor in the response corresponds to a factor in the user\_factor\_profile. Please ensure that every factor has a result.
        \item Analyze the match between the <user\_factor\_profile> and <response\_to\_evaluate> for EACH factor.
        \item Assign a score of 0, 1, or 2. \textbf{Important: The `score` field must be a raw INTEGER type (int), do not use strings.}
        \item Provide a brief reasoning for your score.
        \item Output the result in strict JSON format.
    \end{enumerate}
    \end{tcolorbox}
\caption{Random-Factor Evaluation Prompt.}
\label{fig:Random Evaluation Prompt}
\end{figure*}

\begin{table*}[t!]
\centering
\small
\renewcommand{\arraystretch}{1.2}
\setlength{\tabcolsep}{0pt}
\begin{tabular*}{\textwidth}{@{\extracolsep{\fill}} l ccc ccc ccc ccc @{}}
\toprule
\multirow{2}{*}{\textbf{Method}} & \multicolumn{3}{c}{\textbf{Eng. \& Tools}} & \multicolumn{3}{c}{\textbf{Science}} & \multicolumn{3}{c}{\textbf{Lifestyle}} & \multicolumn{3}{c}{\textbf{Leisure}} \\
\cmidrule(lr){2-4} \cmidrule(lr){5-7} \cmidrule(lr){8-10} \cmidrule(lr){11-13}
 & Acc$\uparrow$ & Tie$\downarrow$ & Mar.$\uparrow$ & Acc$\uparrow$ & Tie$\downarrow$ & Mar.$\uparrow$ & Acc$\uparrow$ & Tie$\downarrow$ & Mar.$\uparrow$ & Acc$\uparrow$ & Tie$\downarrow$ & Mar.$\uparrow$ \\
\midrule
Direct       & 12.15 & 80.67 & 0.026 & 18.86 & 75.00 & 0.065 & 10.17 & 84.75 & 0.027 & 11.90 & 81.75 & 0.028 \\
CoT          & 26.00 & 60.31 & 0.061 & 31.35 & 55.16 & 0.091 & 26.39 & 61.74 & 0.079 & 29.55 & 59.85 & 0.106 \\
Jaccard      & 40.28 & --    & --    & 42.32 & --    & --    & 39.23 & --    & --    & 42.06 & --    & --    \\
Incl. Coeff. & 34.07 & --    & --    & 26.54 & --    & --    & 27.60 & --    & --    & 28.17 & --    & --    \\
Random       & 46.18 & 44.00 & 0.184 & 50.22 & 41.67 & 0.210 & 47.94 & 40.68 & 0.169 & 52.78 & 35.32 & 0.233 \\
\textbf{ours} & \textbf{55.09} & \textbf{28.24} & \textbf{0.189} & \textbf{55.04} & \textbf{30.92} & \textbf{0.211} & \textbf{56.17} & \textbf{24.21} & \textbf{0.180} & \textbf{57.14} & \textbf{24.21} & \textbf{0.216} \\
\bottomrule
\end{tabular*}
\caption{Comparison of Evaluation Methodologies. Arrows ($\uparrow$/$\downarrow$) indicate whether higher or lower values are better. Values are in percentages (\%) except for Margin.}
\label{tab:Comparison of Evaluation Methodologies_appendix}
\end{table*}

\section{Generalizability of the Factors}\label{sec:Generalizability of the Factors_appendix}
Table\ref{tab:UPGCQA} show the result of UPGC-QA. we select the open-source Qwen3-8B (Fast-Thinking mode) \cite{yang2025qwen3} and the proprietary GPT-4o-mini \cite{hurst2024gpt} to serve as backbone models. 
Figure \ref{fig:System Factors Prompt} shows the factor prompt that is integrated into the system prompt. We adopt the experimental settings and evaluation metrics from the original papers to assess the effectiveness of our proposed factors. 

\begin{table}[t]
\centering
\small
\renewcommand{\arraystretch}{1.2} 
\begin{tabular}{l cccc}
\toprule
\textbf{Method} & \textbf{A\&E} & \textbf{L\&P} & \textbf{S\&C} & \textbf{Avg.} \\
\midrule
\multicolumn{5}{c}{\textbf{Qwen3 8B}} \\ 
\midrule
No-Pers.             & 0.3422 & 0.4824 & 0.4824 & 0.4357 \\
RAG-Pers.            & 0.3634 & 0.4700 & 0.4983 & 0.4439 \\
\hspace{3mm}\textit{w/ Factors} & \textbf{0.3743} & \textbf{0.4892} & \textbf{0.5131} & \textbf{0.4589} \\
\midrule
\multicolumn{5}{c}{\textbf{GPT-4o-mini}} \\
\midrule
No-Pers.             & 0.3941 & 0.5162 & 0.5282 & 0.4795 \\
RAG-Pers.            & 0.4182 & 0.4861 & 0.5343 & 0.4795 \\
\hspace{3mm}\textit{w/ Factors} & \textbf{0.4378} & \textbf{0.5234} & \textbf{0.5720} & \textbf{0.5111} \\
\bottomrule
\end{tabular}
\caption{Performance comparison on LaMP-QA. \textbf{A\&E}: Art \& Entertainment, \textbf{L\&P}: Lifestyle \& Personal Development, \textbf{S\&C}: Society \& Culture. ``w/ Factors'' denotes the baseline augmented with our factor prompts.}
\label{tab:lampqa}
\end{table}

\begin{figure}[h!]
    \small
    \centering
    \begin{tcolorbox}[colframe=RoyalBlue, colback=white]
    Your response must incorporate personalization by addressing the following six dimensions:
    \vspace{1mm} 
    \begin{enumerate}
        \item \textbf{Cognitive Trust:} Does the response align with the user's threshold for trust and credibility within this specific domain?
        \item \textbf{Situational Anchoring:} Is the response precisely calibrated to the user's immediate context and specific problem?
        \item \textbf{Schema Consistency:} Does the response integrate coherently with the user's prior knowledge and existing mental models?
        \item \textbf{Cognitive Load Management:} Is the complexity of the response tailored to match the user's cognitive capacity?
        \item \textbf{Metacognitive Scaffolding:} Does the response provide structural support that fosters the user's critical thinking skills?
        \item \textbf{Affective and Motivational Resonance:} Does the response resonate with the user's current emotional state and motivational orientation?
    \end{enumerate}
    \end{tcolorbox}
\caption{Factors Prompt.}
\label{fig:System Factors Prompt}
\end{figure}

\section{Instructions for Human Annotators}\label{sec:Instructions for Human Annotators}
You will be given three pieces of information: (1) a user factor profile, (2) a question, and (3) a candidate response. Your task is to judge how well the response matches the user's preferences on the following six dimensions: Cognitive Trust, Situational Anchoring, Schema Consistency, Cognitive Load Management, Metacognitive Scaffolding, and Affective and Motivational Resonance.

For each dimension, assign one score using a 3-point Likert scale:
\begin{itemize}
    \item 0 (\textit{Mismatch}): the response clearly violates or ignores the user's preference;
    \item 1 (\textit{Partial Match}): the response partially addresses the user's preference, but lacks sufficient depth or completeness;
    \item 2 (\textit{Full Match}): the response is well aligned with the user's preference on this dimension.
\end{itemize}

Please evaluate each sample independently based only on the provided materials. Do not assume any additional user background beyond the given profile. If you are uncertain between two scores, choose the more conservative one.

\section{Factor-level evaluation of personalized QA approaches using \ourM{}}\label{sec:Baselines for the CoPA Benchmark Prompt}
Figure \ref{fig:No-Personalization Prompt} shows the prompt used for No-Personalization. Figure \ref{fig:Time-Personalization Prompt} shows the prompt used for Time-Personalization. Figure \ref{fig:RAG-Personalization Prompt} shows the prompt used for RAG-Personalization.

\subsection{Detailed factor-level analysis}\label{sec:Detailed factor-level analysis}
Tables ~\ref{tab:Detailed factor-level analysis of Engineering&Tools} --\ref{tab:Detailed factor-level analysis of Leisure&Fandom} show the detailed factor-level analysis.
\begin{table*}[htbp]
    \centering
    \resizebox{\textwidth}{!}{%
        \begin{tabular}{lcccccc}
            \toprule
            \textbf{Method} 
            & \makecell[c]{\textbf{Cognitive} \\ \textbf{Trust}} 
            & \makecell[c]{\textbf{Situational} \\ \textbf{Anchoring}} 
            & \makecell[c]{\textbf{Schema} \\ \textbf{Consistency}} 
            & \makecell[c]{\textbf{Cognitive Load} \\ \textbf{Management}} 
            & \makecell[c]{\textbf{Metacognitive} \\ \textbf{Scaffolding}} 
            & \makecell[c]{\textbf{Affective \& Motiv.} \\ \textbf{Resonance}} \\
            \midrule
            
            \multicolumn{7}{c}{\textbf{GPT-4o-mini}} \\
            \midrule
            No-Personalization      & 0.5972 & 0.6348 & 0.6574 & 0.7737 & 0.4213 & 0.6997 \\
            Time-Personalization    & 0.6562 & 0.6985 & 0.7153 & 0.8183 & 0.4641 & 0.7622 \\
            RaG-Personalization     & 0.6719 & 0.7020 & 0.7338 & 0.8200 & 0.4797 & 0.7714 \\
            Profile-Personalization & 0.6743 & 0.7101 & 0.7344 & 0.8163 & 0.4835 & 0.7832 \\
            
            \midrule
            
            \multicolumn{7}{c}{\textbf{Qwen3-8b}} \\
            \midrule
            No-Personalization      & 0.6441 & 0.6765 & 0.6973 & 0.7726 & 0.4589 & 0.7535 \\
            Time-Personalization    & 0.6933 & 0.7297 & 0.7564 & 0.8403 & 0.5000 & 0.7934 \\
            RaG-Personalization     & 0.6921 & 0.7367 & 0.7569 & 0.8299 & 0.5111 & 0.7922 \\
            Profile-Personalization & 0.7135 & 0.7442 & 0.7841 & 0.8200 & 0.5272 & 0.8154 \\
            \bottomrule
        \end{tabular}%
    }
    \caption{Detailed factor-level analysis of Engineering\&Tools}
    \label{tab:Detailed factor-level analysis of Engineering&Tools}
\end{table*}

\begin{table*}[htbp]
    \centering
    \resizebox{\textwidth}{!}{%
        \begin{tabular}{lcccccc}
            \toprule
            \textbf{Method} 
            & \makecell[c]{\textbf{Cognitive} \\ \textbf{Trust}} 
            & \makecell[c]{\textbf{Situational} \\ \textbf{Anchoring}} 
            & \makecell[c]{\textbf{Schema} \\ \textbf{Consistency}} 
            & \makecell[c]{\textbf{Cognitive Load} \\ \textbf{Management}} 
            & \makecell[c]{\textbf{Metacognitive} \\ \textbf{Scaffolding}} 
            & \makecell[c]{\textbf{Affective \& Motiv.} \\ \textbf{Resonance}} \\
            \midrule
            
            \multicolumn{7}{c}{\textbf{GPT-4o-mini}} \\
            \midrule
            No-Personalization      & 0.4452 & 0.5197 & 0.5022 & 0.5855 & 0.2621 & 0.5055 \\
            Time-Personalization    & 0.5044 & 0.5822 & 0.5680 & 0.6524 & 0.3279 & 0.5768 \\
            RaG-Personalization     & 0.5033 & 0.5789 & 0.5548 & 0.6579 & 0.3311 & 0.5746 \\
            Profile-Personalization & 0.5513 & 0.6321 & 0.6285 & 0.7122 & 0.3982 & 0.6254 \\
            \midrule
            
            \multicolumn{7}{c}{\textbf{Qwen3-8b}} \\
            \midrule
            No-Personalization      & 0.5132 & 0.5888 & 0.5844 & 0.6568 & 0.3279 & 0.5746 \\
            Time-Personalization    & 0.5296 & 0.6283 & 0.6118 & 0.6941 & 0.3487 & 0.6020 \\
            RaG-Personalization     & 0.5340 & 0.6305 & 0.6173 & 0.6787 & 0.3596 & 0.6140 \\
            Profile-Personalization & 0.6151 & 0.6974 & 0.6842 & 0.7303 & 0.4254 & 0.6667 \\
            \bottomrule
        \end{tabular}%
    }
    \caption{Detailed factor-level analysis of Science\&Theory}
    \label{tab:Detailed factor-level analysis of Science&Theory}
\end{table*}

\begin{table*}[htbp]
    \centering
    \resizebox{\textwidth}{!}{%
        \begin{tabular}{lcccccc}
            \toprule
            \textbf{Method} 
            & \makecell[c]{\textbf{Cognitive} \\ \textbf{Trust}} 
            & \makecell[c]{\textbf{Situational} \\ \textbf{Anchoring}} 
            & \makecell[c]{\textbf{Schema} \\ \textbf{Consistency}} 
            & \makecell[c]{\textbf{Cognitive Load} \\ \textbf{Management}} 
            & \makecell[c]{\textbf{Metacognitive} \\ \textbf{Scaffolding}} 
            & \makecell[c]{\textbf{Affective \& Motiv.} \\ \textbf{Resonance}} \\
            \midrule
            
            \multicolumn{7}{c}{\textbf{GPT-4o-mini}} \\
            \midrule
            No-Personalization      & 0.4855 & 0.5896 & 0.5581 & 0.6961 & 0.2603 & 0.6186 \\
            Time-Personalization    & 0.4891 & 0.6235 & 0.5835 & 0.7276 & 0.2906 & 0.6404 \\
            RaG-Personalization     & 0.5024 & 0.6356 & 0.5920 & 0.7470 & 0.3063 & 0.6659 \\
            Profile-Personalization & 0.5139 & 0.7018 & 0.6577 & 0.7827 & 0.3766 & 0.7014 \\
            \midrule
            
            \multicolumn{7}{c}{\textbf{Qwen3-8b}} \\
            \midrule
            No-Personalization      & 0.5182 & 0.6465 & 0.6029 & 0.7312 & 0.3220 & 0.6610 \\
            Time-Personalization    & 0.5835 & 0.6743 & 0.6501 & 0.7688 & 0.3644 & 0.6973 \\
            RaG-Personalization     & 0.5678 & 0.6780 & 0.6392 & 0.7676 & 0.3656 & 0.6852 \\
            Profile-Personalization & 0.6235 & 0.7300 & 0.7094 & 0.7930 & 0.4262 & 0.7554 \\
            \bottomrule
        \end{tabular}%
    }
    \caption{Detailed factor-level analysis of Lifestyle\&Society}
    \label{tab:Detailed factor-level analysis of Lifestyle&Society}
\end{table*}

\begin{table*}[htbp]
    \centering
    \resizebox{\textwidth}{!}{%
        \begin{tabular}{lcccccc}
            \toprule
            \textbf{Method} 
            & \makecell[c]{\textbf{Cognitive} \\ \textbf{Trust}} 
            & \makecell[c]{\textbf{Situational} \\ \textbf{Anchoring}} 
            & \makecell[c]{\textbf{Schema} \\ \textbf{Consistency}} 
            & \makecell[c]{\textbf{Cognitive Load} \\ \textbf{Management}} 
            & \makecell[c]{\textbf{Metacognitive} \\ \textbf{Scaffolding}} 
            & \makecell[c]{\textbf{Affective \& Motiv.} \\ \textbf{Resonance}} \\
            \midrule
            
            \multicolumn{7}{c}{\textbf{GPT-4o-mini}} \\
            \midrule
            No-Personalization      & 0.4405 & 0.5357 & 0.5357 & 0.6845 & 0.3274 & 0.5575 \\
            Time-Personalization    & 0.4901 & 0.5873 & 0.5813 & 0.7044 & 0.3512 & 0.5952 \\
            RaG-Personalization     & 0.4802 & 0.5675 & 0.5833 & 0.6944 & 0.3611 & 0.5972 \\
            Profile-Personalization & 0.5310 & 0.5922 & 0.6117 & 0.7053 & 0.3566 & 0.6348 \\
            \midrule
            
            \multicolumn{7}{c}{\textbf{Qwen3-8b}} \\
            \midrule
            No-Personalization      & 0.4841 & 0.5774 & 0.5635 & 0.6825 & 0.3492 & 0.5853 \\
            Time-Personalization    & 0.5020 & 0.6032 & 0.5972 & 0.7004 & 0.3710 & 0.6091 \\
            RaG-Personalization     & 0.5198 & 0.6052 & 0.6032 & 0.7143 & 0.3611 & 0.6290 \\
            Profile-Personalization & 0.5496 & 0.6349 & 0.6369 & 0.7044 & 0.4127 & 0.6627 \\
            \bottomrule
        \end{tabular}%
    }
    \caption{Detailed factor-level analysis of Leisure\&Fandom}
    \label{tab:Detailed factor-level analysis of Leisure&Fandom}
\end{table*}

\begin{figure}[h!]
    \small
    \centering
    \begin{tcolorbox}[colframe=RoyalBlue, colback=white]
    \textbf{System:}
    \\
    Role: You are an intelligent assistant skilled in teaching.
    Your task is to generate personalized answers to user questions.
    \\
    \vspace{1mm}
    
    \textbf{User:}
    \\
    User's Question:
    \\
    \{question\}
    \\
    \vspace{1mm}
    
    The response should be in JSON format:
    \\
    \{\\
      \hspace*{1em}"answer": "",\\
      \hspace*{1em}"reasoning": "Brief explanation..."\\
    \}
    \\
    Requirements:
    \\
    1. The answer needs to be accurate.
    \\
    2. Reasoning is a brief explanation of how you arrived at the answer above.
    \\
    3. Ensure the output adheres to the specified format.
    \end{tcolorbox}
    \caption{No-Personalization Prompt.}
    \label{fig:No-Personalization Prompt}
\end{figure}

\begin{figure}[h!]
    \small
    \centering
    \begin{tcolorbox}[colframe=RoyalBlue, colback=white]
    \textbf{System:}
    \\
    Role: You are an intelligent assistant skilled in teaching.
    Your task is to generate responses tailored to the user's individual understanding, based on their question history.
    \\
    \vspace{1mm}
    
    \textbf{User:}
    \\
    User's Historical Questions:
    \\
    \{Historical\_questions\}
    \\
    User's Current Question:
    \\
    \{Current\_question\}
    \\
    \vspace{1mm}
    
    The response should be in JSON format:
    \\
    \{\\
      \hspace*{1em}"answer": "",\\
      \hspace*{1em}"reasoning": "Brief explanation..."\\
    \}
    \\
    Requirements:
    \\
    1. The answer needs to be accurate.
    \\
    2. Reasoning is a brief explanation of how you arrived at the answer above.
    \\
    3. Ensure the output adheres to the specified format.
    \end{tcolorbox}
    \caption{Time-Personalization Prompt.}
    \label{fig:Time-Personalization Prompt}
\end{figure}

\begin{figure}[h!]
    \small
    \centering
    \begin{tcolorbox}[colframe=RoyalBlue, colback=white]
    \textbf{System:}
    \\
    Role: You are an intelligent assistant skilled in teaching.
    Your task is to generate responses tailored to the user's individual understanding, based on their question history.
    \\
    \vspace{1mm}
    
    \textbf{User:}
    \\
    User's Historical Questions:
    \\
    \{Historical\_questions\}
    \\
    User's Current Question:
    \\
    \{Current\_question\}
    \\
    \vspace{1mm}
    
    The response should be in JSON format:
    \\
    \{\\
      \hspace*{1em}"answer": "",\\
      \hspace*{1em}"reasoning": "Brief explanation..."\\
    \}
    \\
    Requirements:
    \\
    1. The answer needs to be accurate.
    \\
    2. Reasoning is a brief explanation of how you arrived at the answer above.
    \\
    3. Ensure the output adheres to the specified format.
    \end{tcolorbox}
    \caption{RAG-Personalization Prompt.}
    \label{fig:RAG-Personalization Prompt}
\end{figure}
\subsection{Impact of User History Context Length (K) on Model Performance}\label{sec:Impact of User History Context Length (K) on Model Performance}
\begin{figure}[h!]
\centering
    \includegraphics[width=1\columnwidth]{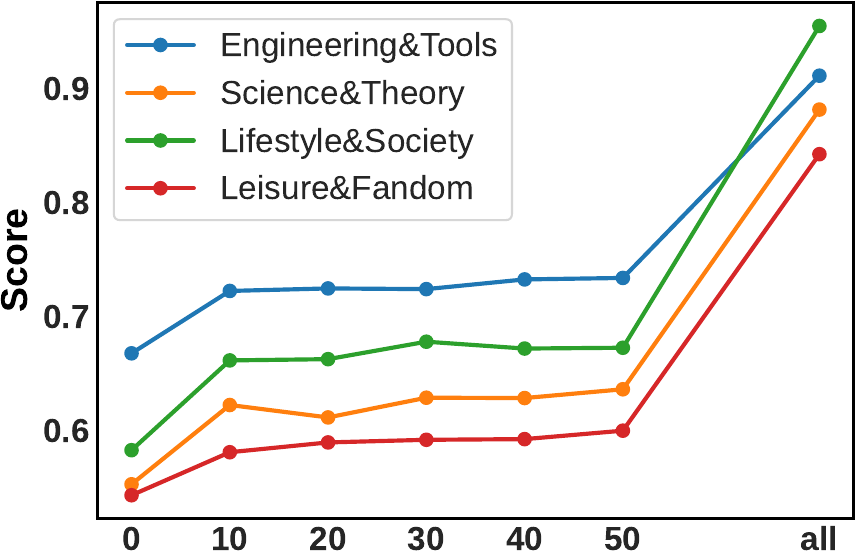}
    \caption{Impact of User History Context Length (K) on Model Performance}
    \label{fig:K_Line chart}
\end{figure}

Figure \ref{fig:K_Line chart} illustrates the specific impact of incorporating varying lengths of user query history (K) on model performance across four sub-domains. Overall, the experimental results demonstrate a significant positive correlation between the inclusion of user history and model performance, with the full history (k = all) yielding the most substantial gains. However, it is important to note that increasing K inevitably leads to higher inference time costs. This trade-off provides valuable insights and directions for future research.

\subsection{Profile-Personalization}\label{sec:Profile-Personalization Prompt}
Figures~\ref{fig:Domain Extract Prompt}--\ref{fig:Answer Generation Prompt} show the prompts used for Profile-Personalization.
\begin{figure}[h!]
    \small
    \centering
    \begin{tcolorbox}[colframe=RoyalBlue, colback=white]
    \textbf{System:}
    \\
    Role: You are an intelligent assistant.
    \\
    \vspace{1mm}
    
    \textbf{User:}
    \\
    question:
    \\
    \{question\}
    \\
    Please tell me which domain of the question is related to.
    \\
    Requirements:
    \\
    1. Please provide your decision in JSON format, following this structure:
    \\
    \{\\
      \hspace*{1em}"domain": "A summarization of which domain this question is related to" (if you are unable to summarize it, please set this value to "None"),\\
      \hspace*{1em}"reasoning": "briefly explain your reasoning for the summarization"\\
    \}
    \\
    2. Please ensure 'domain' is a single word.
    \\
    3. The "reasoning" has no word limits.
    \\
    4. Do not provide any other text outside the JSON string.
    \end{tcolorbox}
    \caption{Domain Extract Prompt.}
    \label{fig:Domain Extract Prompt}
\end{figure}

\begin{figure}[h!]
    \small
    \centering
    \begin{tcolorbox}[colframe=RoyalBlue, colback=white]
    \textbf{System:}
    \\
    You are an expert in \{domain\}. Please tell me the profile the user who asked the question in \{domain\}.
    \\
    \vspace{1mm}
    
    \textbf{User:}
    \\
    question:\{question\}
    \\
    Requirements:
    \\
    1. Please provide your summary in JSON format, following this structure:
    \\
    \{\\
      \hspace*{1em}"profile": "A Summary of the user's profile in this domain"(starting with "This user"),\\
      \hspace*{1em}"reasoning": "briefly explain your reasoning for the summarization"\\
    \}
    \\
    2. Please ensure that the "profile" is accurate.
    \\
    3. Do not provide any other text outside the JSON string.
    \end{tcolorbox}
    \caption{Domain Profile Extract Prompt.}
    \label{fig:Domain Profile Extract Prompt}
\end{figure}

\begin{figure}[h!]
    \small
    \centering
    \begin{tcolorbox}[colframe=RoyalBlue, colback=white]
    \textbf{System:}
    \\
    You are an expert in \{domain\}. Please generate a new user profile based on the user's historical profile and the current profile.
    \\
    \vspace{1mm}
    
    \textbf{User:}
    \\
    history user profile:\{history\}
    \\
    current user profile:\{current\}
    \\
    \vspace{1mm}
    
    Requirements:
    \\
    1. Please provide your output in JSON format, following this structure:
    \\
    \{\\
      \hspace*{1em}"profile": "A Summary of the user's profile in this domain"(starting with "This user"),\\
      \hspace*{1em}"reasoning": "briefly explain your reasoning for the summarization"\\
    \}
    \\
    2. Please ensure that the "profile" is accurate.
    \\
    3. Do not provide any other text outside the JSON string.
    \end{tcolorbox}
    \caption{Domain Profile Synthesize Prompt.}
    \label{fig:Domain Profile Synthesize Prompt}
\end{figure}

\begin{figure}[h!]
    \small
    \centering
    \begin{tcolorbox}[colframe=RoyalBlue, colback=white]
    \textbf{System:}
    \\
    You are an intelligent assistant. Please summarize the user profile based on the information about the user in various domains.
    \\
    \vspace{1mm}
    
    \textbf{User:}
    \\
    user global profile: \{global\_profile\}
    \\
    user domain profile: \{domain\_profile\}
    \\
    \vspace{1mm}
    
    Requirements:
    \\
    1. Please provide your output in JSON format, following this structure:
    \\
    \{\\
      \hspace*{1em}"profile": "A Summary of user profile" (starting with "This user"),\\
      \hspace*{1em}"reasoning": "briefly explain your reasoning for the summarization"\\
    \}
    \\
    2. Please ensure that the "profile" is comprehensive and accurate.
    \\
    3. Do not provide any other text outside the JSON string.
    \end{tcolorbox}
    \caption{Global Profile Generate Prompt.}
    \label{fig:Global Profile Generate Prompt}
\end{figure}

\begin{figure}[h!]
    \small
    \centering
    \begin{tcolorbox}[colframe=RoyalBlue, colback=white]
    \textbf{System:}
    \\
    Role: You are an intelligent assistant skilled in teaching.
    Your task is to generate personalized answers to user questions.
    \\
    \vspace{1mm}
    
    \textbf{User:}
    \\
    User's Profile:
    \\
    \{user\_profile\}
    \\
    User's Current Question:
    \\
    \{Current\_question\}
    \\
    \vspace{1mm}
    
    The response should be in JSON format:
    \\
    \{\\
      \hspace*{1em}"answer": "",\\
      \hspace*{1em}"reasoning": "Brief explanation..."\\
    \}
    \\
    Requirements:
    \\
    1. The answer needs to be accurate.
    \\
    2. Reasoning is a brief explanation of how you arrived at the answer above.
    \\
    3. Ensure the output adheres to the specified format.
    \end{tcolorbox}
    \caption{Answer Generation Prompt.}
    \label{fig:Answer Generation Prompt}
\end{figure}

\section{License}
The code and benchmark resources for \ourM{} will be released under the Apache License 2.0.
This license permits use, modification, and redistribution subject to the terms of the license.

\section{AI Assistance Usage}
Gemini\footnote{\url{https://gemini.google.com/}} was employed as a linguistic assistant to improve readability. Specifically, we used the tool to refine the preliminary drafts of certain sections. All AI-assisted text underwent rigorous manual review and editing to ensure precision and coherence before finalization.


\end{document}